\documentclass[a4paper,fleqn]{cas-sc}

\usepackage[authoryear]{natbib}
\usepackage{array}

\providecommand{\checkmark}{\surd}
\DeclareRobustCommand{\yes}{\ensuremath{\checkmark}}
\newcolumntype{Y}[1]{>{\raggedright\arraybackslash}p{#1\linewidth}}

\begin{document}
\let\WriteBookmarks\relax
\def\floatpagepagefraction{1}
\def\textpagefraction{.001}

\shorttitle{Humanoid robots for safety-critical switching}
\shortauthors{S. Liu et al.}

\title[mode=title]{Humanoid robots for safety-critical electrical switching: A perception-informed runtime constraint filter}

\author[1]{Songyang Liu}
\ead{liusongyang@ufl.edu}

\author[2]{Yu Hou}
\ead{yu.hou@charlotte.edu}
\author[3]{Yuqing Hu}
\ead{yfh5204@psu.edu}

\author[1]{Shuai Li}
\cormark[1]
\ead{shuai.li@ufl.edu}

\affiliation[1]{organization={Department of Civil and Coastal Engineering, University of Florida},
                country={United States}}
\affiliation[2]{organization={Albert School of Construction, William States Lee College of Engineering, University of North Carolina at Charlotte}, country={United States}}
\affiliation[3]{organization={Department of Architectural Engineering, The Pennsylvania State University}, country={United States}}

\cortext[cor1]{Corresponding author}

\begin{abstract}
Humanoid robots can reduce personnel exposure in electrical switching, but safe control is challenging because the tool must reach a target switch next to similar devices, where an imprecise operator command can contact a non-target switch. This study proposes a runtime safety filter that checks each operator command during virtual-reality teleoperation of a humanoid robot. The filter integrates a scene model estimated from the robot's head camera and a target-conditioned control barrier function quadratic program that keeps the tool away from the non-target switch while passing safe commands unchanged. In real-world experiments on a Unitree G1 humanoid robot, with six rounds at each of two standing positions, target operations succeed in all ten rounds with an unchanged target and no non-target actuation. A simulation study achieves 9/9 safe completions, compared with 5/9 for raw teleoperation, demonstrating the feasibility of the method for humanoid switching operations.

\end{abstract}



\maketitle

\section{Introduction}
\label{sec:Introduction}

\begin{figure}
    \centering
    \includegraphics[width=.5\linewidth]{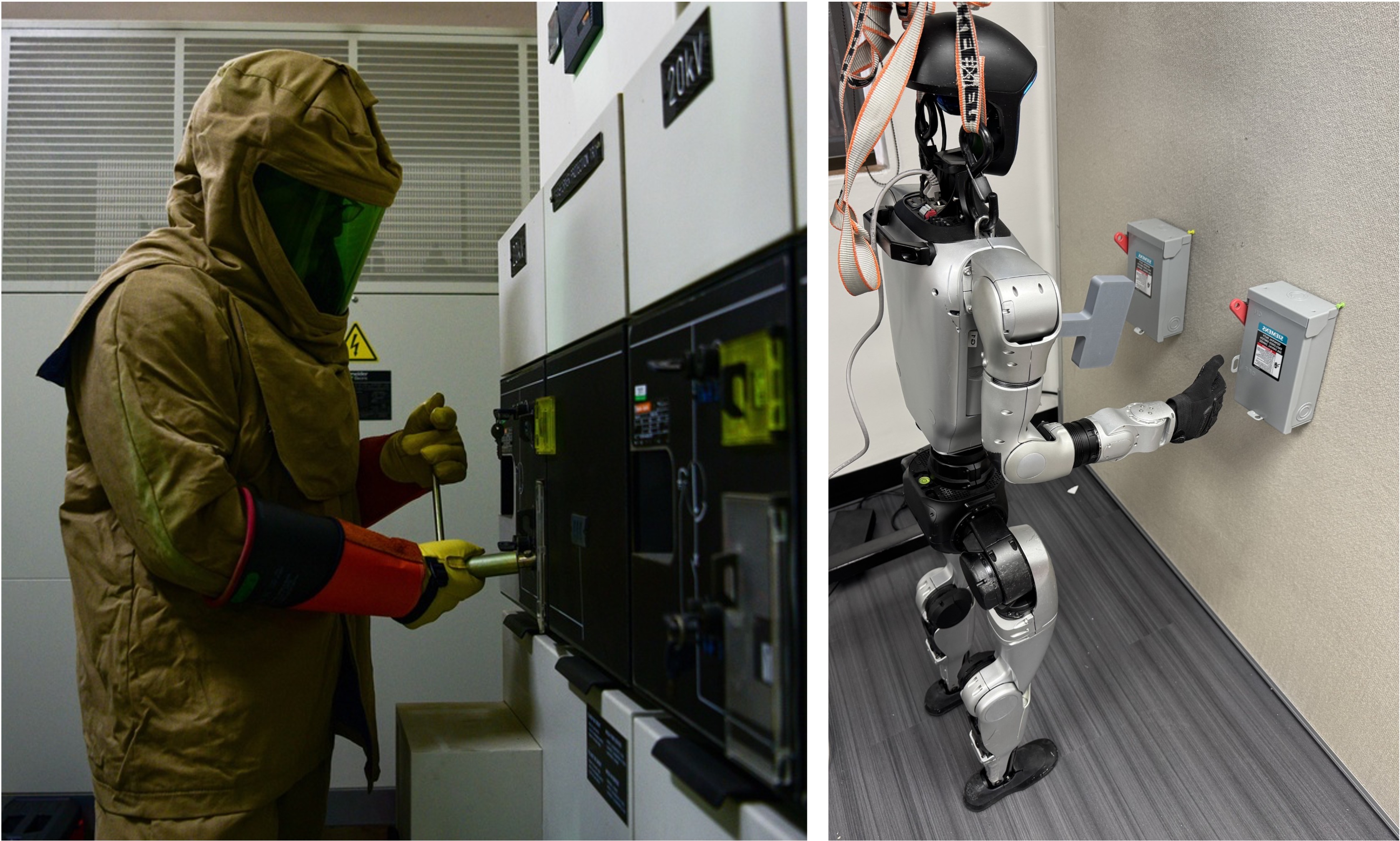}
    \caption{Switch operation by a person and by a humanoid robot. Left: a worker in arc-flash protective clothing operates switchgear~\citep{dvids2022cemirt}. Right: Unitree G1 humanoid robot operates  switches in our lab.}
    \label{fig:intro}
\end{figure}

\begin{figure}
    \centering
    \includegraphics[width=.8\linewidth]{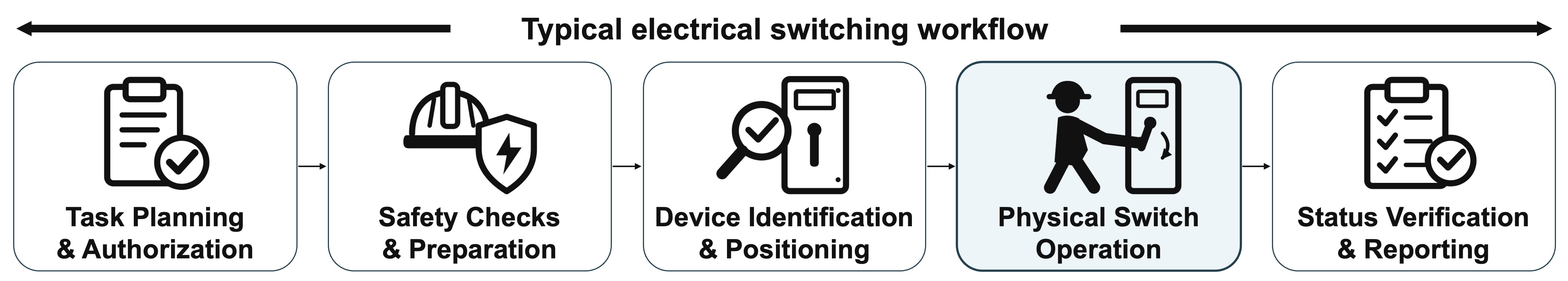}
    \caption{A typical electrical switching workflow, based on utility switching procedures \citep{snopud2026switching,prpa2021switching}. This work addresses the highlighted step, physical switch operation.}
    \label{fig:procedure}
\end{figure}

Electrical substations are critical infrastructure in modern power systems, where
transmission and distribution equipment must be switched, isolated, restored, and
verified to maintain grid reliability. Although a switching action may appear to
be a simple physical operation, such as opening or closing a disconnect switch,
breaker, or switchgear handle, power-system switching is a safety-critical
procedure performed near equipment that is or may be energized. Occupational Safety and Health Administration (OSHA) requires
safety-related work practices to prevent electric shock and other injuries when
employees work on or near energized equipment or circuits; exposed live parts
shall generally be de-energized before work is performed unless de-energization
is infeasible or introduces additional hazards, and only qualified persons may
work on energized equipment under appropriate precautions
\citep{osha1910_333}. These requirements indicate that switching is not merely
a manipulation problem, but a hazardous intervention task involving electrical
risk, procedural control, and human safety (Fig.~\ref{fig:intro}, left). Utility switching manuals further show that switching must be executed as an
authorized and verified sequence rather than as an isolated device actuation.
The Snohomish County Public Utility District switching and clearance manual
establishes the authority, responsibilities, and procedures for system operation,
clearance, and switching, and states that system operators issue orders to open
and close switches, place or remove safety cards, and de-energize or clear
equipment \citep{snopud2026switching}. Similarly, the Platte River Power
Authority power-system switching procedure requires the switchman to carry the
switching program, verify the device identification nameplate and location,
recheck the correct sequence, operate the device, and verify that the device has
moved to the desired position \citep{prpa2021switching}. These procedures
highlight that electrical safety switching requires not only physical manipulation,
but also perception, target identification, communication, operator supervision,
and post-operation verification (Fig.~\ref{fig:procedure}). This work addresses the physical operation step of this workflow.

Robots have increasingly been introduced into substations to reduce human
exposure and improve inspection efficiency. Existing substation robots are
commonly designed for autonomous patrol, visual inspection, thermal imaging,
gauge reading, and anomaly detection. Recent work on substation inspection
robots focuses on navigation, multi-sensor fusion, obstacle perception, and
route-based inspection in complex substation environments
\citep{yang2025substation}. Industrial deployments of mobile and quadruped
robots similarly emphasize routine visual and thermal inspection of
transformers, breakers, and other assets \citep{avangrid2024spot}. These
systems are valuable for monitoring and condition assessment, but they do not
fully address contact-rich switching operations that require human-scale
reachability, accurate end-effector alignment, controlled force interaction, and
verification of a mechanical state change on legacy equipment. In contrast,
device-specific remote switching or racking systems can reduce human exposure
for particular switchgear, but they often require pre-installed hardware,
compatible equipment, or task-specific adaptation.

This motivates the use of humanoid robots for safety switching. The goal
is not to claim that humanoids are universally superior to wheeled robots,
quadrupeds, fixed manipulators, or specialized remote actuators. Rather, the
motivation is that many electrical facilities and operating procedures are
designed around human morphology and human tool use. Switch handles, access
holes, cabinet layouts, doors, stairs, narrow passages, and hot-stick-style
interfaces implicitly assume human reach, posture, bimanual operation, and
visual confirmation. Humanoid robots provide a human-scale body, arms, mobile
base, and whole-body redundancy that can potentially reuse existing
human-centered infrastructure with less environmental modification. This
advantage is particularly relevant when the target facility is not redesigned
for automation and when the robot must act as a remote physical avatar in a
workspace originally intended for human operators. A fixed or rail-mounted arm can operate a known device reliably, but it must be installed at each panel and reaches only the devices within its workspace. A mobile manipulator avoids this installation, yet its base and arm height are usually chosen for a particular task, and stairs, doorways and raised thresholds in older facilities can restrict where it can go. A humanoid can, in principle, walk to a panel, stand where a worker would stand and use the same handles and tools. The method proposed in this paper, however, does not depend on the humanoid form: it filters the arm command and could be placed on another manipulator. What the humanoid adds to the safety problem is that the robot stands at a different place each time it approaches a panel, so the position of the neighboring devices relative to the arm changes from one visit to the next and cannot be fixed by a one-off calibration. For this reason, the constraints in this work are rebuilt from the robot's own head camera at its current standing position.

Using robots can reduce human exposure to electrical hazards, but it does not eliminate the hazards themselves. Human workers near energized equipment rely on arc-rated clothing and insulated tools, and a robot near the same equipment would also need protection, since its sensors, motors and onboard computers can be damaged by an arc or disturbed by electromagnetic fields. This study focuses on the motion of the robot relative to neighboring devices, and protection of the robot itself is discussed in our Discussion section.
Fig.~\ref{fig:intro} (right) shows the humanoid used in this study at a pair of neighboring safety switches.


Safety-switch operation is not only a precision-control problem. Its main safety challenge comes from intentional contact with a selected switch in a confined workspace, next to neighboring devices of the same type, with limited operator visibility and imperfect or delayed teleoperation commands. The robot must be allowed to contact the target switch with its designated tool or finger, while its arms, hands, torso and tool stay clear of the cabinet and neighboring switches, and which device counts as the target changes with the operator's selection. A safety layer for this task therefore has to correct unsafe operator commands continuously, adapt its constraints to the selected target, and stop inward motion toward a non-target device in a way that an unchanged operator command cannot restart.


In this paper, we present a runtime safety filter for virtual-reality (VR) teleoperation of a humanoid robot on neighboring safety switches. Here, runtime means that our filter operates inside the control loop while the robot is being teleoperated: in every control cycle, it checks the operator's command against the scene model before the command is sent to the robot, and modifies the command only when it would bring the tool too close to a non-target device. This differs from approaches that plan a safe motion in advance or train safe behavior offline. An operator commands the robot's arms through a PICO headset, and an existing retargeting and whole-body control stack generates the nominal command. Instead of replacing this stack with a task-specific learned policy, we insert a minimum-intervention filter between its upper-body command and the robot. A scene model estimated from the robot's head red-green-blue-depth (RGB-D) camera defines the constraints, and the operator's choice of target decides which objects they apply to. Safe commands pass through unchanged, and at the boundary the arm returns to its initial pose and waits for re-authorization. No reinforcement- or imitation-learning training is required. We evaluate the filter on a physical Unitree G1 operating two wall-mounted safety switches. Separately, in a MuJoCo pushbutton task, we extend it with part-region contact permissions and phase-dependent motion directions, and compare it with raw teleoperation and two ablations.

We make the following contributions:
\begin{itemize}
    \item We propose a target-conditioned runtime safety filter for humanoid teleoperation, which uses an onboard RGB-D scene model to keep the tool away from non-target devices while leaving the selected target reachable, passes safe commands unchanged, and adds part-region and phase constraints in simulation.

    \item We conduct experiments on a physical Unitree G1 humanoid robot (10/12 operator-reported target operations, no non-target actuation) and in a paired simulation study (9/9 safe completions versus 5/9 for raw teleoperation), to demonstrate the effectiveness of our proposed method.
\end{itemize}

\section{Related Work}
\label{sec:related-work}

\subsection{Robots for Electrical Substations}
\label{sec:rw-substations}

Surveys of substation robots cover navigation and inspection technologies \citep{allan2014substations,lu2017substation}, and a technical brochure of the International Council on Large Electric Systems (CIGRE) reports that more than 1,000 outdoor patrol robots are in full operation worldwide \citep{cigre2020robotics}. Robots that physically act on equipment are less common. The same brochure describes a teleoperated patrol robot whose arm can operate a disconnect switch, and a breaker-racking prototype \citep{cigre2020robotics}. \citet{lin2024grounding} automate the hanging and removal of grounding rods, reporting 19 successes in 20 trials on a practical robot. \citet{jia2025switchgear} use an end-to-end learned perception network so that an autonomous robot can operate switchgear buttons, knobs and handcarts. \citet{zhao2025teleoperation} teleoperate a Unitree H1 on an electrical box with buttons, breakers and rotational switches, comparing teleoperation devices and manipulator controllers.

Inspection robots mainly monitor equipment rather than operate it, while the robots of Lin et al.\ and Jia et al.\ execute specific tasks autonomously, without an operator in the loop. Zhao et al.\ keep an operator in the loop on a humanoid, but their controller, based on a quadratic program (QP), enforces joint limits, and the paper does not describe environment-aware constraints around the box. Our work examines this gap in teleoperation: a runtime filter constrains the tool of a teleoperated humanoid against a neighboring switch while leaving the selected target unconstrained.

\subsection{Humanoid Teleoperation}
\label{sec:rw-teleoperation}

VR interfaces have made it practical to operate humanoids through human motion. Open-TeleVision gives the operator immersive stereo feedback while mirroring arm and hand motion on the robot \citep{cheng2024opentelevision}. OmniH2O uses kinematic pose as a control interface and supports real-time whole-body teleoperation through a VR headset \citep{he2024omnih2o}. TWIST2 captures whole-body operator motion with a PICO headset and motion trackers, without a motion-capture system, for humanoid data collection \citep{ze2025twist2}. SONIC scales up motion tracking and accepts VR teleoperation among its inputs \citep{luo2026sonic}.

These works focus on tracking, immersion and data collection rather than on scene-aware constraints. As \citet{gadde2026recal} note, most whole-body controllers are blind to scene geometry, so imperfect teleoperation targets can cause collisions. In switch operation, the same imperfection can bring the tool into a neighboring device. We therefore keep an existing teleoperation stack unchanged and add a safety layer between its upper-body command and the robot, which passes safe commands through unmodified.

\subsection{Safety Filters and Perception-Informed Constraints}
\label{sec:rw-safety-perception}

Control barrier functions (CBFs) let a quadratic program minimally modify a nominal command to keep the system in a safe set \citep{ames2019cbf}. Table~\ref{tab:related-work-comparison} compares the closest work that adds such filters or scene-aware constraints to robot motion. On humanoids, p-SSA filters live teleoperation of a G1 using obstacle locations sensed by an Apple Vision Pro \citep{chen2025pssa}, and ConstrainedMimic enforces runtime constraints on pre-trained tracking policies on a G1, in simulation and on hardware \citep{morton2026constrainedmimic}. ARMOR uses arm-mounted time-of-flight sensors on a GR1 to train a collision-avoidance policy \citep{kim2024armor}, and RECAL adapts whole-body commands on a Digit V3 from egocentric point clouds \citep{gadde2026recal}. On non-humanoid arms, \citet{selvaggio2016vision} generate virtual fixtures on the fly from an eye-in-hand stereo camera, \citet{brunke2025semantic} enforce semantic constraints with CBFs using a prebuilt scene map, and VLPSA builds safety functions online for a CBF-QP filter and clears pick-and-place targets from its occupancy map \citep{wilkinson2026vlpsa}.

\begin{table}[width=\linewidth,cols=7,pos=t]
\caption{Comparison with existing work. NR: not reported.}
\label{tab:related-work-comparison}
\footnotesize
\begin{tabular*}{\tblwidth}{@{\extracolsep{\fill}}lcccccc@{}}
\toprule
\textbf{Method} & \textbf{Robot} & \shortstack{\textbf{Live}\\\textbf{teleop.}} & \shortstack{\textbf{Online}\\\textbf{geometry}} & \shortstack{\textbf{Onboard}\\\textbf{sensing}} & \shortstack{\textbf{Contact}\\\textbf{permission}} & \shortstack{\textbf{Phase}\\\textbf{directions}} \\
\midrule
ARMOR \citep{kim2024armor}                & GR1        & NR          & \yes        & \yes     & NR     & NR \\
p-SSA \citep{chen2025pssa}                & G1         & \yes        & \yes  & $\times$ & NR     & NR \\
\citet{brunke2025semantic}                & Franka FR3 & \yes        & Prebuilt    & NR       & NR     & NR \\
ConstrainedMimic \citep{morton2026constrainedmimic} & G1 & Sim.\ only & Assumed     & $\times$ & NR     & NR \\
RECAL \citep{gadde2026recal}              & Digit V3   & Loco.\ only & \yes        & \yes     & NR     & NR \\
VLPSA \citep{wilkinson2026vlpsa}          & Franka FR3 & Demos only  & \yes        & NR       & Object & NR \\
\midrule
\textbf{Ours} & \textbf{G1} & \textbf{\yes} & \textbf{Per scene} & \textbf{\yes} & \textbf{Part region} & \textbf{\yes} \\
\bottomrule
\end{tabular*}

\vspace{3pt}
\end{table}

As Table~\ref{tab:related-work-comparison} shows, these works leave a specific combination open. Selvaggio et al., Brunke et al.\ and VLPSA do not use a humanoid. Live teleoperation is not reported for ARMOR, covers only locomotion in RECAL and only simulation in ConstrainedMimic, and is used in VLPSA only to collect demonstrations. p-SSA does not use robot-mounted sensing, and the authors of ConstrainedMimic assume environment geometry and leave onboard perception to future work. Only VLPSA reports clearing a target at the object level. RECAL allows intended contact surfaces to be removed from its input point cloud, but its authors note that it treats perceived geometry as forbidden contact. In contrast, our hardware study combines live humanoid teleoperation, constraints from a robot-mounted RGB-D camera, and object-level exclusion of an operator-selected target. Part-region permissions and phase-dependent motion directions, which no compared work reports, are studied separately in simulation, on a different task with ground-truth geometry.

\section{Methodology}

\subsection{System Overview}
\label{sec:method-overview}

\begin{figure}
    \centering
    \includegraphics[width=\linewidth]{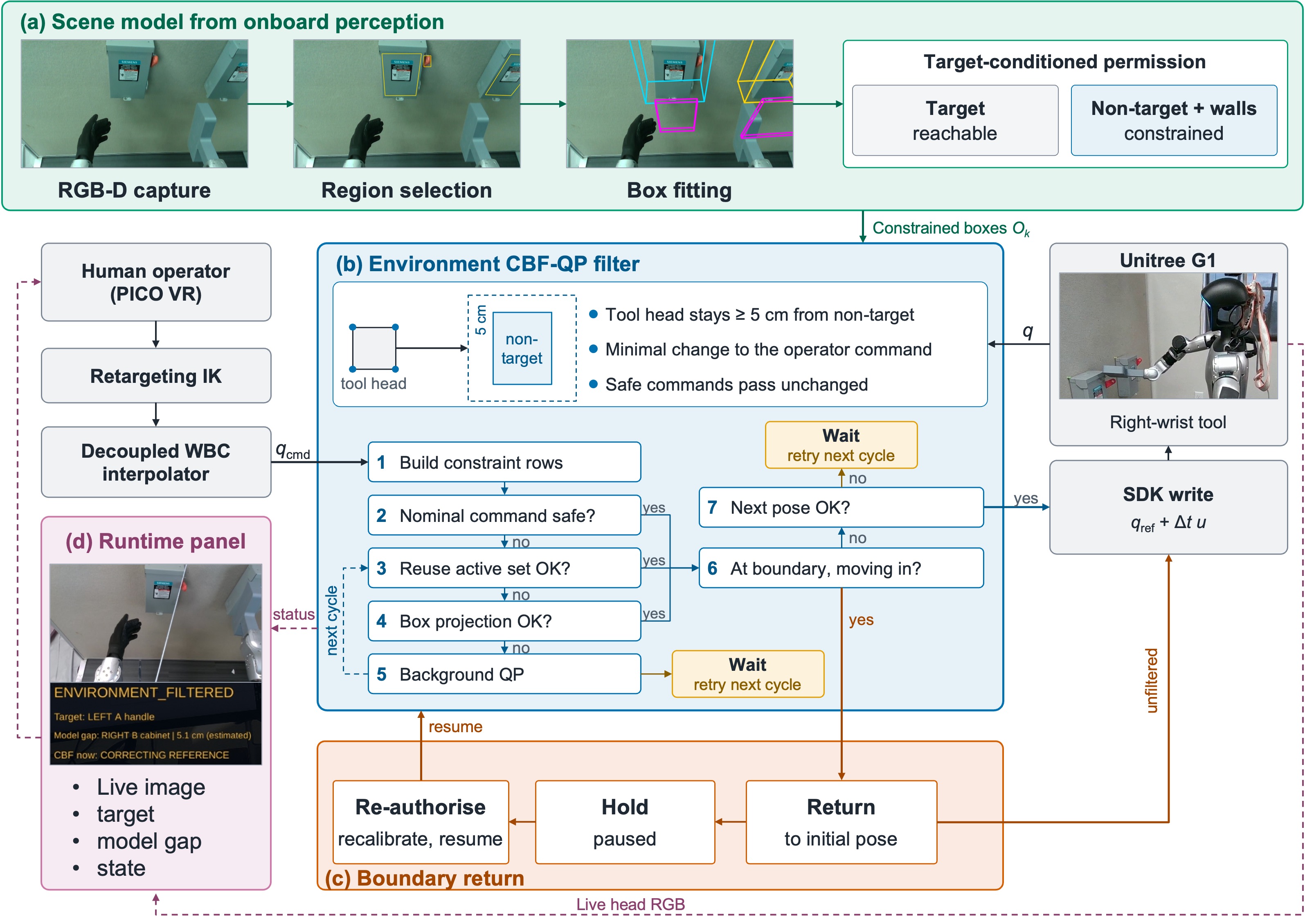}
    \caption{Methodology overview.}
    \label{fig:method_overview}
\end{figure}

Fig.~\ref{fig:method_overview} shows our proposed method overview. A human operator wearing a PICO 4 Ultra headset commands the arms through XRoboToolkit \citep{zhao2025xrobotoolkit}. The retargeting inverse kinematics and the upper-body interpolator of the decoupled whole-body controller convert the operator's hand poses into a nominal joint command $q_{\mathrm{cmd}}$ at a nominal 50~Hz, limited to 1.5~rad/s. Locomotion is not the focus of this study. The robot walks to its standing position before each configuration and then stands under a balance policy, while only the arms are teleoperated, so that the arm motion next to the neighboring switches, which is the step addressed by the filter, is isolated from walking and balance. A rigid 3D-printed tool is mounted on the right wrist.

We keep this teleoperation stack unchanged and add three components around it. (a) Before teleoperation, a scene model is built from the robot's head RGB-D camera: regions selected in the current frame are fitted as oriented boxes, the operator-selected target is excluded so that it remains reachable, and the non-target assembly and the walls become the constrained boxes $O_k$. (b) In every control cycle, an environment CBF-QP filter between $q_{\mathrm{cmd}}$ and the Unitree software development kit (SDK) write constrains the tool head to stay at least 5~cm from these boxes while changing the operator's command as little as possible, and a command that is already safe passes unchanged. To ensure the control loop does not have to wait for the solver to complete its calculations, the filter performs a rapid validation check before executing the background QP optimization. If no feasible command can be found, or if the subsequent pose fails the validation check, the robotic arm remains in its current reference state. (c) If the tool is already at this boundary and the command still moves toward a constrained object, a boundary return takes over: the arm returns to its initial pose outside the filter, holds, and resumes only after recalibration and an explicit operator command. (d) An operator panel in the headset shows the operator the live head image, the selected target, the model gap and the filter state.

\subsection{Scene Model from Onboard Perception}
\label{sec:method-scene}

The scene model is built once, before control, from the robot's head-mounted RGB-D camera while the robot stands still. The camera records about 900 color and depth frames at 640$\times$360 pixels and 30~fps. Four kinds of image region mark the objects of interest: the visible front of each cabinet, each red handle, and two wall regions below the cabinets. The regions are selected per scene by viewing the current frame and are entered through per-session scripts.

From these regions, we build the model in four steps, detailed below: it fits a plane to each cabinet-front and wall region, turns each plane into an oriented box, removes the target assembly so that the tool can reach it, and binds the remaining boxes into the model used during control.

\textbf{Plane fitting.} Each pixel $(u, v)$ inside a region with depth $z$ between 0.1 and 3~m is back-projected to a 3D point $\mathbf{p} = z\,\big((u - c_x)/f_x,\ (v - c_y)/f_y,\ 1\big)^{\top}$ in the camera frame, where $f_x, f_y, c_x, c_y$ are the camera intrinsics. A plane $\{\mathbf{p} : \mathbf{n}^{\top}\mathbf{p} = o\}$ is then fitted to these points, where $\mathbf{n}$ is the unit normal, chosen to point towards the camera, and $o$ is the plane offset. The fit is repeated four times, each time on an updated set $\mathcal{K}$ of trusted points (the inliers), which starts as all points. In each round,
\begin{equation}
\begin{aligned}
\mathbf{n} &= \text{direction of least spread of } \{\mathbf{p}_i - \bar{\mathbf{p}}\}_{i \in \mathcal{K}}, \qquad
o = \operatorname*{median}_{i \in \mathcal{K}} \mathbf{n}^{\top}\mathbf{p}_i, \\
r_i &= \lvert \mathbf{n}^{\top}\mathbf{p}_i - o \rvert, \qquad
\mathcal{K} \leftarrow \{\, i : r_i \le \max(4~\mathrm{mm},\ 3\sigma) \,\},
\end{aligned}
\label{eq:planefit}
\end{equation}
where $\bar{\mathbf{p}}$ is the mean of the inliers. The direction of least spread is the last right singular vector in a singular value decomposition (SVD) of the centered inliers, which gives the least-squares plane. The residual $r_i$ is the distance of point $i$ from the plane, and it is computed for all points, so a point dropped in one round can return in the next. The noise level $\sigma$ is estimated robustly as $\sigma = 1.4826\,\mathrm{MAD}$, where the median absolute deviation (MAD) is the median of $\lvert r_i - \operatorname{median}_{j \in \mathcal{K}} r_j \rvert$ over the inliers. The factor 1.4826 makes $\sigma$ equal to the standard deviation for Gaussian noise. Points more than three noise levels from the plane, such as depth outliers, are therefore dropped, and the 4~mm floor keeps the threshold from collapsing on a very flat surface. Wall regions are fitted differently in two respects. The initial inlier set comes from random sample consensus (RANSAC): in each of 160 trials, a plane is drawn through three random points, and the plane with the most points within 9.07~mm is kept. All four rounds then use this fixed 9.07~mm band instead of $3\sigma$.

\textbf{Box axes.} Each fitted plane defines a right-handed box frame. Let $\mathbf{g}$ be the world vertical expressed in the camera frame, obtained from the robot kinematics at capture. The axes are
\begin{equation}
\mathbf{e}_{\mathrm{h}} = \frac{\mathbf{g} - (\mathbf{n}^{\top}\mathbf{g})\,\mathbf{n}}{\lVert \mathbf{g} - (\mathbf{n}^{\top}\mathbf{g})\,\mathbf{n} \rVert}, \qquad
\mathbf{e}_{\mathrm{w}} = \mathbf{e}_{\mathrm{h}} \times \mathbf{n}, \qquad
\mathbf{R} = [\,\mathbf{e}_{\mathrm{w}}\ \ \mathbf{e}_{\mathrm{h}}\ \ \mathbf{n}\,],
\label{eq:boxaxes}
\end{equation}
so the height axis $\mathbf{e}_{\mathrm{h}}$ is the vertical projected onto the plane, the depth axis is the plane normal, and the width axis $\mathbf{e}_{\mathrm{w}}$ is perpendicular to both. Each inlier is expressed in this frame as $(w_i, h_i, d_i)^{\top} = \mathbf{R}^{\top}\mathbf{p}_i$.

\textbf{Cabinet boxes.} A cabinet box has fixed half-extents $\mathbf{a} = (a_{\mathrm{w}}, a_{\mathrm{h}}, a_{\mathrm{d}}) = (0.105,\ 0.13,\ 0.055)$~m, half of a declared upper-bound size of 0.21$\times$0.26$\times$0.11~m (width, height, depth). Its center is
\begin{equation}
\mathbf{c} = \mathbf{R}
\begin{pmatrix}
\tfrac{1}{2}\big(\min_i w_i + \max_i w_i\big) \\
\min_i h_i - 0.01 + a_{\mathrm{h}} \\
o + 0.02 - a_{\mathrm{d}}
\end{pmatrix},
\label{eq:boxcentre}
\end{equation}
with the minima and maxima taken over the inliers, in meters. In words, the box is centered between the leftmost and rightmost inliers, its bottom face sits 1~cm below the lowest visible inlier, and its front face sits 2~cm in front of the fitted plane, towards the camera, so that protruding parts such as the handle stay inside it. The box then extends 0.11~m behind its front face.

\textbf{Wall boxes.} Each wall becomes a thin box in the frame of its own plane (Eq.~\ref{eq:boxaxes}). Its width and height span both the inliers and the selected region, whose corners are projected onto the fitted plane, and its depth is $o \pm 9.07$~mm.

\textbf{Target and constrained objects.} Before a trial, the operator selects the left or right switch as the target. The tool must be able to reach the target, so the whole target assembly, including its cabinet, is removed from the constraints. The three remaining boxes, the non-target assembly and the two walls, are the constrained objects $O_k$, $k = 1, 2, 3$. A target can be selected only while the system is paused with no active trial, and a new calibration is required before following resumes.

\textbf{Binding.} Before binding, the model must pass a check added during the study (Section~\ref{sec:exp-setup}): the model distance between the tool at the return pose (Section~\ref{sec:method-return}) and the non-target box must be at least 10~cm. An operator command then binds the model, which cannot change afterwards.

\subsection{Environment CBF-QP Filter}
\label{sec:method-cbf}
Section~\ref{sec:method-scene} ends with a fixed model: three boxes $O_k$ that the tool must not approach. This section describes how that model is used during control. At every control step, the native interpolator turns the operator's motion into a joint command for the arms. Before the command is sent, the filter predicts how it would change the model gap between the tool head and each box. Because the boxes do not change after binding and the arm pose comes from the joint encoders, this prediction needs no new image. If the command keeps the tool at least a safety margin $d$ away from every box, and approaches that margin only gradually, it is sent unchanged. Otherwise, the filter replaces it with the closest command that meets these conditions. We express this rule as a CBF on each model gap and compute the corrected command by solving a small QP. The rest of this section defines, in order, the nominal command, the barrier, the constraint rows built from it, the joint limits, and the QP.

The decision variable $u \in \mathbb{R}^{14}$ is the velocity of the 14 arm joints. With $q_{\mathrm{cmd}}$ the interpolator output, $q_{\mathrm{ref}}$ the last sent command and $\Delta t$ the actual sample interval of the native interpolator (nominally 0.02~s), the nominal velocity is
\begin{equation}
u_{\mathrm{nom}} = (q_{\mathrm{cmd}} - q_{\mathrm{ref}})/\Delta t .
\label{eq:nominal}
\end{equation}

The tool head is a rigid box with eight corners $p_v(q)$. For a constrained box $O_k$, an exact box-to-box separating axis test gives the closest exterior unit normal $n_k$. With $s_k$ the largest value of $n_k^{\top} c$ over the corners $c$ of $O_k$, the barrier for corner $v$ is
\begin{equation}
h_{kv}(q) = n_k^{\top} p_v(q) - s_k - d ,
\label{eq:barrier}
\end{equation}
where $d = 0.05$~m. Using all eight corners ensures that a parallel face contact cannot hide the rotation of other corners. With the corner Jacobian $J_v(q)$ over the 14 arm joints, each barrier gives one constraint row
\begin{equation}
-n_k^{\top} J_v(q)\, u \le \gamma \max(h_{kv}, 0)/\Delta t ,
\label{eq:rows}
\end{equation}
with $\gamma = 0.35$, an effective gain $\gamma/\Delta t \approx 17.5~\mathrm{s}^{-1}$ at 50~Hz. For $h_{kv} \ge 0$, the row allows $h_{kv}$ to shrink by at most the fraction $\gamma$ per step, so that to first order $h_{kv}^{+} \ge (1-\gamma)\, h_{kv}$. For $h_{kv} < 0$, the right side is zero: the row forbids going deeper but does not push the tool out. Rows are built at the encoder pose and at the last-sent reference pose, which gives 48 environment rows for three objects and eight corners. Native bounds add 28 rows,
\begin{equation}
\max\!\left(-\bar{u},\, \frac{q_{\min} - q_{\mathrm{ref}}}{\Delta t}\right) \le u \le \min\!\left(\bar{u},\, \frac{q_{\max} - q_{\mathrm{ref}}}{\Delta t}\right),
\label{eq:native}
\end{equation}
with $\bar{u} = 1.5$~rad/s and $q_{\min}$, $q_{\max}$ the joint position limits. Stacking all 76 rows as $G u \le b$, the filter solves
\begin{equation}
\min_{u} \; \frac{1}{2}\| u - u_{\mathrm{nom}} \|^2 \quad \mathrm{s.t.} \quad G u \le b ,
\label{eq:qp}
\end{equation}
using DAQP \citep{arnstrom2022daqp} through qp solvers, with primal and dual tolerances of $10^{-9}$. The sent command is $q_{\mathrm{ref}} + \Delta t\, u$. A command that already satisfies all rows is the exact optimum and passes through unchanged.

\subsection{Non-Blocking Scheduling}
\label{sec:method-scheduling}

Equation~(\ref{eq:qp}) defines which command the filter should send, but not how to obtain it within one control cycle. The loop runs at a nominal 50~Hz, so a command is due about every 20~ms, while the time of a QP solve varies from cycle to cycle. If the loop called the solver and waited for it in every cycle, a slow solve would delay the command to the robot. 

We utilize two properties of the problem to achieve a non-blocking schedule. First, when the tool is far from every box, the nominal command already satisfies all rows, so it is the exact optimum of Eq.~(\ref{eq:qp}) and no solver is needed. Second, the state changes little between consecutive cycles, so the rows that were active in the last solution are likely to be active again, and the new solution can be recovered from them by one small linear solve. The scheduler therefore tries these cheap candidates first, checks each against the optimality conditions of the current QP, and passes the full QP to a background worker only when they fail. A further issue is that the rows of Eq.~(\ref{eq:rows}) are linearized at the current pose, so a command that satisfies them can still shrink a barrier more than allowed at the pose the robot actually reaches. Every candidate is therefore checked again at its next pose before it is sent. If no candidate is ready, or none passes this check, the arm holds its last reference instead of waiting for the solver. 

In each cycle, it applies the following steps in order:
\begin{enumerate}
    \item Build the constraint rows from the current state and the fixed model. If the geometry check fails, hold.
    \item Accept $u_{\mathrm{nom}}$ if $G u_{\mathrm{nom}} \le b$ within a round-off tolerance of at most $10^{-12}$, without calling a solver.
    \item Otherwise, reuse the active set $\mathcal{A}$ of the last accepted solution. Solve $(G_{\mathcal{A}} G_{\mathcal{A}}^{\top})\lambda = G_{\mathcal{A}} u_{\mathrm{nom}} - b_{\mathcal{A}}$ and set $u = u_{\mathrm{nom}} - G_{\mathcal{A}}^{\top}\lambda$. The result is accepted only if it satisfies the Karush-Kuhn-Tucker (KKT) conditions of the current problem: primal and dual feasibility within $10^{-8}$, stationarity and complementarity within $2 \times 10^{-7}$, and a condition number of $G_{\mathcal{A}} G_{\mathcal{A}}^{\top}$ of at most $10^{10}$.
    \item Otherwise, if only native bounds are violated, clip $u_{\mathrm{nom}}$ to Eq.~\eqref{eq:native} and accept the result under the same KKT check against the full problem.
    \item Otherwise, submit the problem to a single background worker that keeps only the latest problem, and hold the last reference for this cycle. A running solve is allowed to finish, and a result whose problem has been superseded is discarded. A finished result is not applied directly: it is cached and enters step 3 in a later cycle.
\end{enumerate}
Once a candidate exists, the boundary check of Section~\ref{sec:method-return} runs. If it does not trigger, the candidate is checked at step scales $\alpha \in \{1, 1/2, 1/4, 1/8\}$, with the trial command $q_{\mathrm{ref}} + \alpha\,(q_{\mathrm{c}} - q_{\mathrm{ref}})$ for candidate command $q_{\mathrm{c}}$. Each trial must satisfy the linear rows and, after re-running the kinematics at the next pose $q^{+}$, the nonlinear condition
\begin{equation}
h_{kv}(q^{+}) \ge h_{kv}(q) - \gamma \max(h_{kv}(q), 0)
\label{eq:stepcheck}
\end{equation}
for all rows, and must not leave the tool at the boundary while closing. The first trial that passes is sent. If every scale fails, the arm holds its last reference. We call any of these holds a constraint hold, shown as Wait in Fig.~\ref{fig:method_overview}. The same operator command is retried in the next cycle, and following continues without operator action once a candidate passes.

\subsection{Boundary Return and Re-Authorization}
\label{sec:method-return}

The filter of Sections~\ref{sec:method-cbf} and \ref{sec:method-scheduling} only limits motion. At the boundary, where a barrier reaches zero, the rows forbid further approach but do not move the tool away (Eq.~\ref{eq:rows}). If the operator keeps commanding inward motion there, the filter keeps removing the inward part of each command, or the scheduler keeps holding, and the tool stays at the boundary for as long as the operator pushes. We therefore treat continued inward motion at the boundary as a separate event. Before a candidate is sent (Section~\ref{sec:method-scheduling}), the scheduler runs the boundary check defined below. When the check triggers, control leaves the filter, the arm returns to its initial pose, and teleoperation stays paused until the operator re-authorizes it.

A return starts when, for some constrained object, the exact model gap is at or below $d$ at the encoder or reference pose and the commanded closing speed toward that object at that pose exceeds $10^{-7}$~m/s. The arm then moves from the last sent command to the initial pose of the trial, stored when teleoperation first starts, through the original interpolator at no more than 1.5~rad/s. This pose lies outside the boundary, which the pre-binding check of Section~\ref{sec:method-scene} enforces. On completion, the arm holds, the operator's last target is discarded, and the system stays paused.

\subsection{Operator Interface}
\label{sec:method-interface}

\begin{figure}
    \centering
    \includegraphics[width=0.5\linewidth]{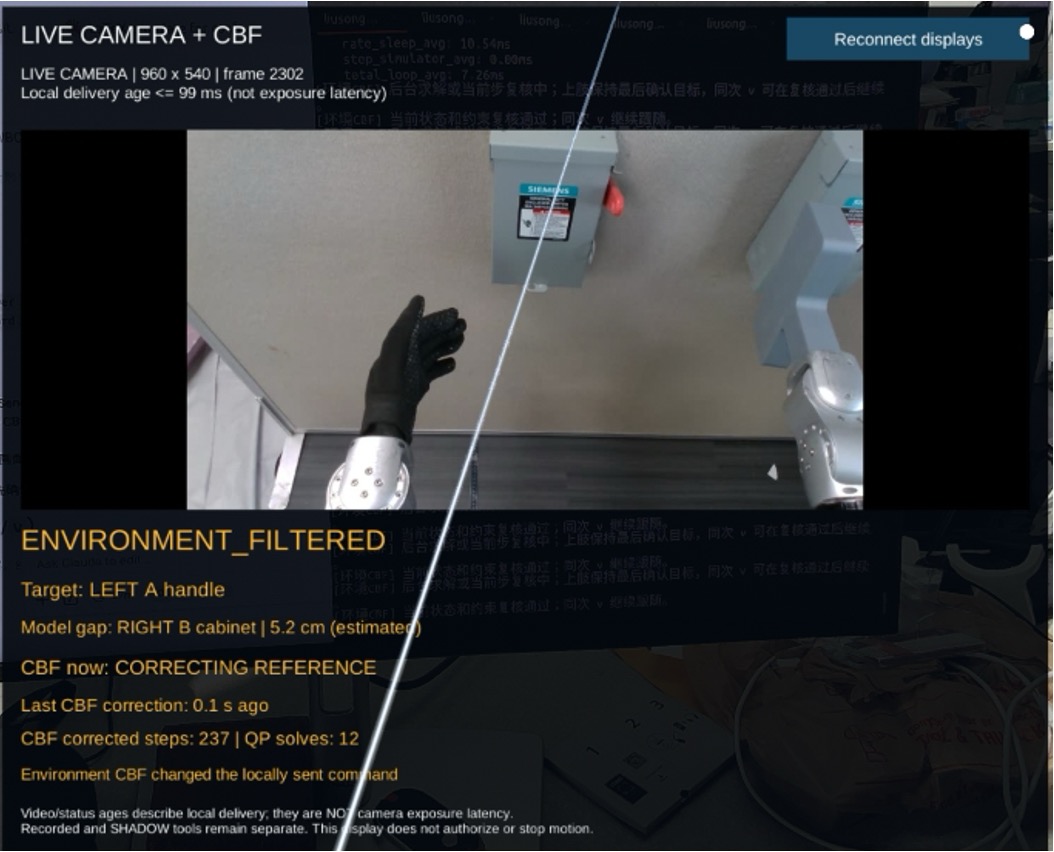}
    \caption{Human operator panel in the PICO headset during filtering, showing the live head-camera image, filter state, selected target, nearest model gap and correction counts.}

    \label{fig:ui}
\end{figure}

A panel in the PICO application (Fig.~\ref{fig:ui}) shows the live head camera image, labeled with its local delivery age, together with the active target, the nearest constrained object with its estimated model gap, and the filter state. The states are paused, following unchanged, environment-filtered, native-limited, returning and waiting. The panel also shows whether the filter is correcting the command, the time since the last correction, and the cumulative numbers of corrected steps and QP solves. Control inputs are separate: calibration, explicit start or resume, pause, return and target selection.

\subsection{Simulation Filter for Part-Region and Phase Constraints}
\label{sec:method-simulation}

Our system of Sections~\ref{sec:method-scene}-\ref{sec:method-interface} grants permission at the object level: the whole target assembly is removed from the constraints, and the same constraints apply throughout the task. It therefore does not specify which part of the robot may touch which part of the target, and it does not restrict the direction of motion while the target is pressed and released. We address these two points with two extensions. Part-region permissions allow contact only between selected parts of the robot and the target, and phase-direction constraints change the allowed motion as the task moves between phases. 

We study both extensions in simulation, on a dedicated task and filter. Fig.~\ref{fig:setup} (a) shows the simulated task. A simulated Unitree G1 presses a guarded spring-return pushbutton with its right index finger, next to two non-target buttons. The button actuates past a 10~mm trigger and stops at 25~mm of travel. The task makes both points concrete: only the index fingertip pad may touch the button, while the rest of the hand stays clear of the guard, and the finger must stay centered while pressing and leave straight outward.

We build the simulation method in the following layers. A base filter, shared by all filtered methods, keeps the robot away from the scene. A phase tracker follows the task through its steps, and the two extensions are added on top of it, since both change with the phase. The section ends with the compared methods, each of which removes one extension.

\textbf{Base filter.} All filtered methods share a kinematic velocity filter of the same form as Section~\ref{sec:method-cbf}. Its decision variable $\dot{q}$ contains 21 joints: the two seven-degree-of-freedom (7-DoF) arms and the seven right-hand finger joints, since contact is made with a finger. It solves $\min_{\dot q} \frac{1}{2}\|\dot q - \dot q_{\mathrm{nom}}\|^2$ with DAQP, subject to joint position limits, joint speed limits of 1.5~rad/s for the arms and 0.4~rad/s for the fingers, and collision rows. A collision row is built for every pair of a collision-enabled robot geom $i$, covering the arms, hands, torso and pelvis, and an environment geom $j$ within 15~cm:
\begin{equation}
-n_{ij}^{\top}\left(J_j - J_i\right)\dot{q} \le k_{ij} \max(d_{ij} - m_{ij}, 0)/\Delta t ,
\label{eq:simcollision}
\end{equation}
where $d_{ij}$ is the distance between the two geoms, $n_{ij}$ the unit normal between their closest points, $J_i$ and $J_j$ the Jacobians at those points, $m_{ij}$ the margin and $k_{ij}$ the gain. Like Eq.~(\ref{eq:rows}), each row lets the distance shrink only gradually towards the margin and forbids further approach once the margin is reached. By default, $m_{ij} = 30$~mm for the cabinet and the target and 60~mm for the non-target buttons, with $k_{ij} = 0.05$. Rows are built at both the measured state and the integrated reference, and the reference may deviate from the measured state by at most 0.35~rad per joint.

\textbf{Phases and transitions.} Both extensions depend on the task phase, so the filter first tracks it. The task has four phases: approach, engage, actuate and retreat. The input requests each phase, and a requested transition is accepted only when its gate holds; otherwise it stays pending. The gates use the index tip offset $e = (e_x, e_y, e_z)$ from the center of the target face, expressed along the button axis ($e_x$, positive inward) and two lateral axes. Engage requires $-90 \le e_x \le 2$~mm, a lateral offset $\|(e_y, e_z)\|$ of at most 25~mm, and an alignment of at least 0.90 between the index pad and the button axis. Actuate requires an index pad gap to the button of at most 2~mm and $|e_y|, |e_z| \le 15$~mm. A new approach, or a change of target, requires a completed retreat, with the pad more than 40~mm from the button and the tip more than 40~mm outside the face.

\textbf{Part-region permission.} The first extension decides which contacts are allowed. During approach, all contacts are protected. From engage until the retreat has separated, a single pair is permitted: the right index fingertip pad with the target button. Its margin becomes $-2$~mm, which allows up to 2~mm of pad compression. All other right-hand parts keep a 5~mm clearance to the target button and its guard, and every other pair keeps its default row. Pairs between the right hand and the target button or guard use gain 0.4 in all phases.

\textbf{Phase-direction constraints.} The second extension decides how the finger may move in each phase. It keeps the tip centered on the button while in contact, stops the press short of the trigger until actuate is accepted, and makes the finger leave straight outward. Let $v = (v_x, v_y, v_z)$ and $\omega$ be the linear and angular velocity of the index tip in the button frame. Table~\ref{tab:sim-phase} lists the rows added in each phase. In engage and actuate, the lateral rows form a centering corridor,
\begin{equation}
\max\!\left(-\bar v_{\ell},\, -\kappa (r_c + e_i)\right) \le v_i \le \min\!\left(\bar v_{\ell},\, \kappa (r_c - e_i)\right), \quad i \in \{y, z\},
\label{eq:corridor}
\end{equation}
with $\bar v_{\ell} = 40$~mm/s, $\kappa = 2~\mathrm{s}^{-1}$ and $r_c = 10$~mm, which keeps the tip inside the 15~mm actuation gate with a 5~mm reserve. During engage, a virtual depth stop $v_x \le \min(25~\mathrm{mm/s},\, 4~\mathrm{s}^{-1}(1~\mathrm{mm} - \delta))$, with $\delta$ the depth of the fingertip past the target face position recorded when engage begins, keeps the press well short of the 10~mm trigger until actuate is accepted. In retreat, the outward cone forbids sideways scraping and rotation while the finger is not moving out. A mechanical-stop row also forbids inward motion once 25~mm of travel is reached.


\textbf{Compared methods.} Our complete method (M3) uses both extensions. Each ablation removes one of them. M1 keeps the phase-direction rows but replaces the part-region permission with a whole-target permission: while permissions are active, it removes every pair between the robot and the target assembly, button and guard. M2 keeps the part-region permission, transition gates, depth stop and mechanical stop, but removes the phase-direction rows. M0 sends the nominal command without a filter.

\begin{table}[width=\linewidth,cols=3,pos=t]
\caption{Phase-direction rows of the simulation filter, in the button frame ($v_x > 0$ inward).}
\label{tab:sim-phase}
\begin{tabular*}{\tblwidth}{@{} Y{0.17} Y{0.38} Y{0.38} @{}}
\toprule
Quantity & Engage and actuate & Retreat (until separated) \\
\midrule
Axial speed $v_x$ & $-25 \le v_x \le 25$~mm/s; $\le 0$ beyond 25~mm travel or outside the 15~mm gate & $-60 \le v_x \le 0$~mm/s \\
Lateral speed $v_y, v_z$ & Centering corridor, Eq.~\eqref{eq:corridor} & $|v_i| \le 30$~mm/s and outward cone $|v_i| \le -0.25\, v_x$ \\
Angular speed $\omega$ & $|\omega_i| \le 0.12$~rad/s & $|\omega_i| \le 0.12$~rad/s and $|\omega_i| \le -2\, v_x$ (rad/s, $v_x$ in m/s) \\
Depth & Virtual stop at 1~mm (engage only) & -- \\
\bottomrule
\end{tabular*}
\end{table}

\section{Experiments and Results}
\label{sec:experiments}

\subsection{Experimental Setup}
\label{sec:exp-setup}

\begin{figure}
    \centering
    \includegraphics[width=.4\linewidth]{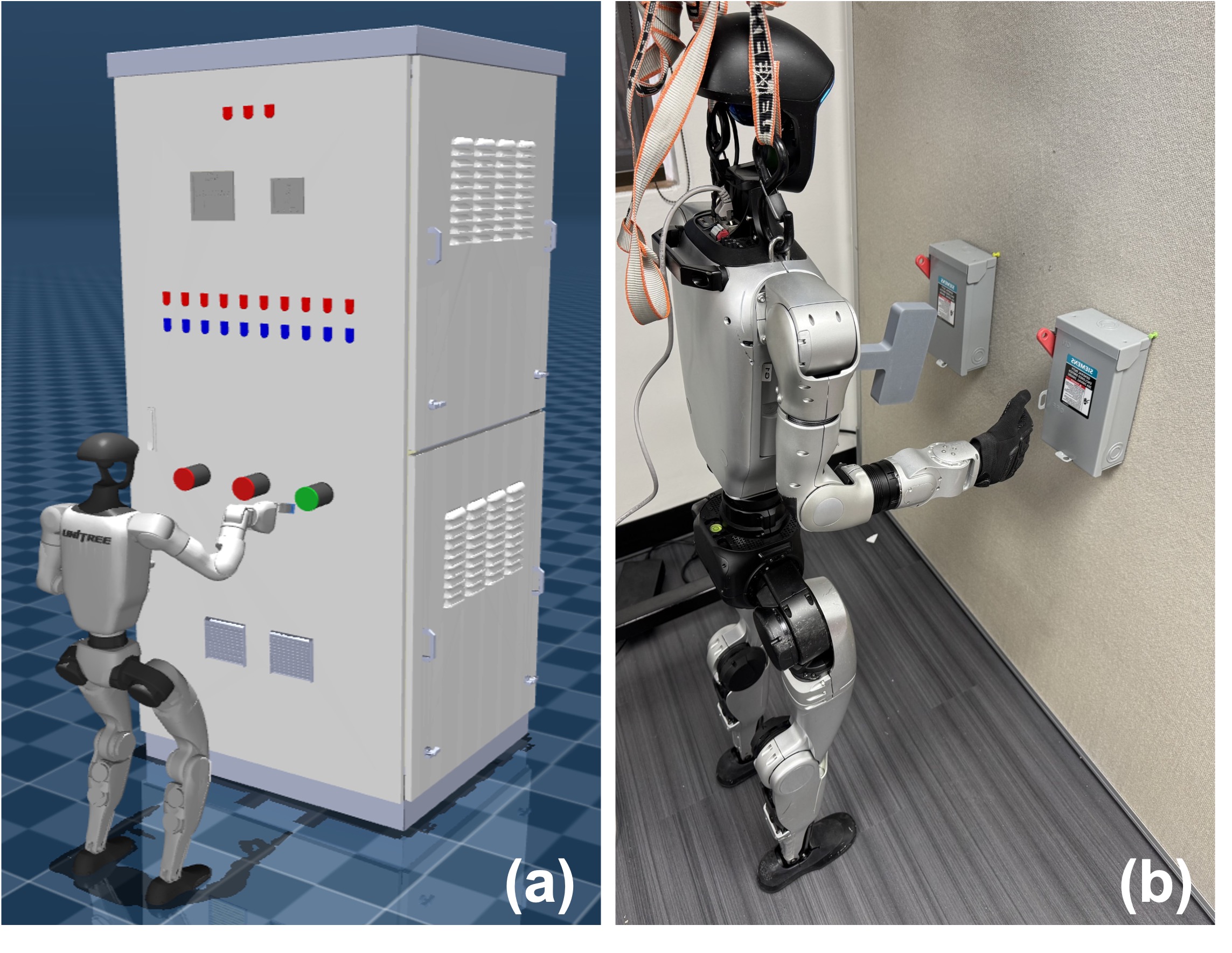}
    \caption{Experimental setups. (a) Simulation: G1 pressing a pushbutton in MuJoCo. (b) Real-world: G1 with a right-wrist tool at two wall-mounted safety switches.}

    \label{fig:setup}
\end{figure}

We conduct two separate studies, in simulation and on real-world hardware, shown in Fig.~\ref{fig:setup}. The real-world study tests our complete system on a physical robot, with the scene model estimated from the robot's own camera. The simulation provides ground-truth contact forces and repeatable inputs, which allow comparison with raw teleoperation and ablations and a test of the part-region and phase-direction extensions, which require a multi-finger hand. The two studies are compared in Section~\ref{sec:disc-sim-real}.

\textbf{Simulation.} The simulation evaluates the task and methods of Section~\ref{sec:method-simulation}. Our simulation study contains three layouts, three inputs and the four methods M0-M3, for 36 paired runs with identical inputs and initial states. The layouts place the cabinet panel and its buttons at three poses relative to the robot: the nominal pose (S1), shifted 30~mm sideways along the panel (S2), and rotated by 5$^{\circ}$ about the vertical axis (S3). In each layout, the filter uses the simulator's geometry of that layout, so S2 and S3 test whether the same input still completes the task when the button is not where the nominal layout puts it.

The inputs are normal (I0); permission-risk (I1), in which a finger pivot risks protected contact; and wrong-direction (I2), which adds lateral and inward motion during engage, actuate and retreat. I1 and I2 therefore probe the part-region permission and the phase-direction constraints, respectively. A run counts as a safe completion if it presses past the trigger in the accepted actuate phase, withdraws, and has no non-permitted contact, instability proxy or timeout. Four further M3 runs compare a fixed old map with updated geometry in S2 and S3.

\textbf{Real-world.} The robot platform is a Unitree G1 with a 3D-printed tool on the right wrist, a head-mounted Intel RealSense camera D435i, and a PICO 4 Ultra headset. The scene has two safety switches mounted on a wall, left A and right B, each with a red side handle.


A human operator conducts six rounds in each of two robot-switch configurations, C1 and C2, giving 12 rounds in total. In the two configurations, the robot stands at different positions relative to the wall and the switches. Within a configuration, the robot stays at the same position, and a single scene model is captured and bound there (Section~\ref{sec:method-scene}), so all six rounds use the same geometry. In rounds 1-5, the target is the same switch, B in C1 and A in C2. In round 6, the operator selects the other switch as the target. Every round consists of the same two steps. The operator first uses the tool to operate the target switch, after which the arm is retracted and recalibrated. The operator then deliberately moves the tool towards the non-target switch until the filter stops further inward motion. The switches are not energized.

\subsection{Simulation Results}
\label{sec:exp-simulation}

Table~\ref{tab:sim-results} summarizes the 36 runs in our simulation study. A run counts as pressed and withdrawn when the button passes the 10~mm trigger in the accepted actuate phase and the finger then withdraws. A safe completion additionally requires no non-permitted contact, instability proxy or timeout. M3 safely completes 9/9 runs, compared with 5/9 for M0 and 6/9 for both M1 and M2. M0 and M1 also press in every run, but four and three of their runs, respectively, have non-permitted contact. M2 fails differently. It does not complete any of its three I1 runs, two of which involve a filter hold, and its one run with non-permitted contact is among these incomplete runs. Across our simulation study, the maximum M3 button travel is 18.74~mm, 8.74~mm past the trigger, and the maximum M3 filter and QP times are 14.59~ms and 0.747~ms. Together, these results show that the two extensions address different failures. Without part-region permission, the task is still completed but with non-permitted contact, and without phase-direction constraints, some runs are not completed. Only with both does every run press, withdraw and avoid non-permitted contact, and M3 achieves this without any filter hold, so its safety does not come from stopping the task.

\begin{table}[pos=t]
\caption{Simulation results over nine paired runs per method (three layouts $\times$ three inputs). }
\label{tab:sim-results}
\begin{tabular*}{\tblwidth}{@{} LCCCCC @{}}
\toprule
Method & Pressed and withdrawn & Safe completion & Non-permitted contact & Incomplete & Held \\
\midrule
M0 Raw          & 9/9 & 5/9 & 4/9 & 0/9 & 0/9 \\
M1 Whole target & 9/9 & 6/9 & 3/9 & 0/9 & 0/9 \\
M2 No direction & 6/9 & 6/9 & 1/9 & 3/9 & 2/9 \\
M3 Complete     & 9/9 & 9/9 & 0/9 & 0/9 & 0/9 \\
\bottomrule
\end{tabular*}
\end{table}

\begin{figure}
    \centering
    \includegraphics[width=0.8\linewidth]{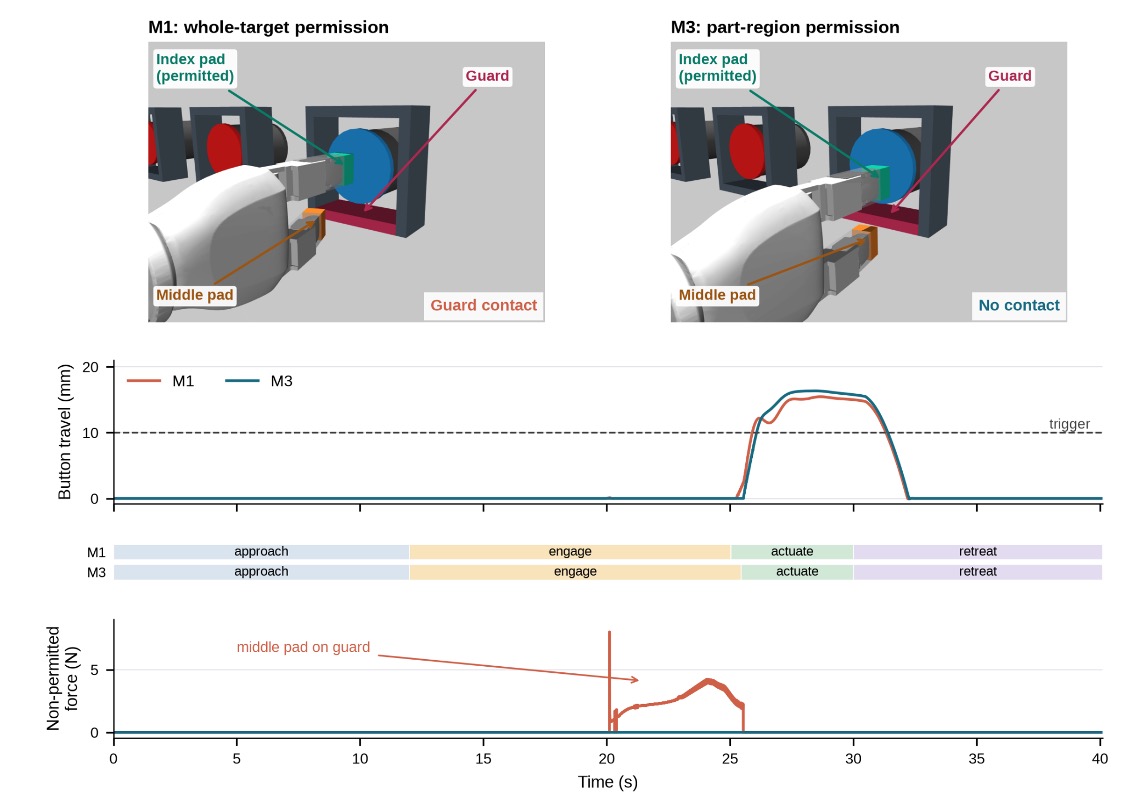}
    \caption{Effect of part-region permission under input I1 in layout S1 in our simulation study. Top: robot state at $t = 24$~s with whole-target (M1) and part-region (M3) permission. Middle: button travel, with the 10~mm trigger dashed. Bottom: summed non-permitted contact force.}
    \label{fig:sim-permission}
\end{figure}

Fig.~\ref{fig:sim-permission} shows how part-region permission separates M1 from M3 under I1. With whole-target permission (M1), nothing keeps the rest of the hand away from the target assembly, and the middle finger pad touches the guard of the target. With part-region permission (M3), the middle pad keeps its clearance. Both methods press past the trigger and withdraw with nearly the same phase timing, so the difference lies only in which parts touch the target. M1 produces non-permitted contact force during engage, with a peak summed force of 8.00, 8.50 and 9.35~N in S1, S2 and S3, and is therefore safe in 0/3 runs. M3 has no non-permitted contact and is safe in 3/3 runs. This indicates that, once contact with the target is allowed, permission at the object level is too coarse: it also lets other parts of the hand touch protected parts of the target, such as the guard. Restricting permission to the intended contact, here the index pad on the button, keeps the rest of the hand clear without preventing the press.

\begin{figure}
    \centering
    \includegraphics[width=0.8\linewidth]{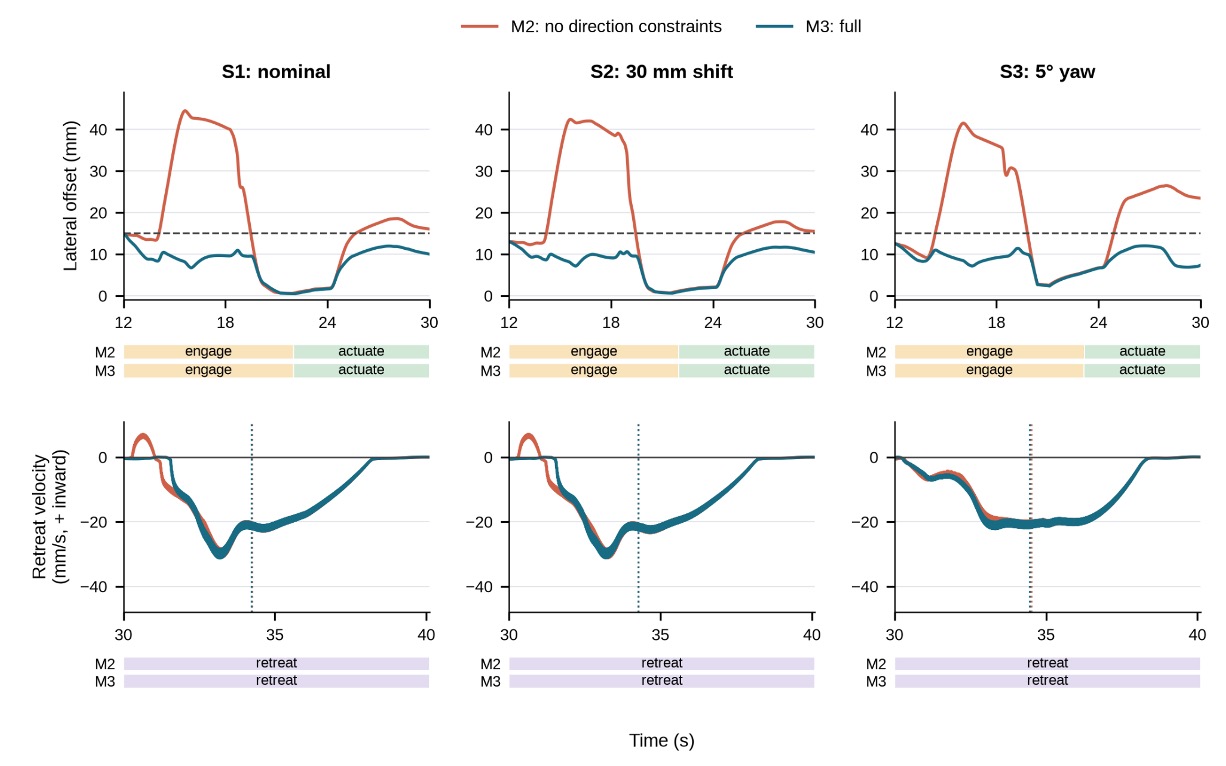}
    \caption{Effect of phase-direction constraints under input I2 in the three layouts. Top: lateral offset of the index tip from the button axis during engage and actuate, with the 15~mm gate dashed. Bottom: axial velocity during retreat, positive inward.}

    \label{fig:sim-directions}
\end{figure}

Fig.~\ref{fig:sim-directions} compares M2 and M3 under I2, where the input adds lateral and inward motion during engage, actuate and retreat. Both methods complete safely in 3/3 runs, so the phase-direction constraints do not change the outcome, but they change the executed motion. Without them (M2), the index tip follows the lateral input and moves well outside the 15~mm gate during engage, whereas the centering corridor keeps the M3 tip close to the button axis. Over engage and actuate, the cumulative lateral path of the index pad falls from 114.53, 117.32 and 116.72~mm with M2 to 48.22, 45.28 and 49.67~mm with M3 in S1, S2 and S3, a reduction of about 57-61\%. At the start of retreat, M2 moves back inward by 3.467~mm in S1 and 3.375~mm in S2, whereas M3 moves inward by only 0.006 and 0.0005~mm. In S3, inward motion during retreat is already small and rises slightly, from 0.016~mm with M2 to 0.021~mm with M3. In this test, the value of the phase-direction constraints therefore lies in how the finger moves rather than in whether the task succeeds. They keep the finger centered on the button while it is in contact and stop it from pushing back in as it withdraws, except in S3, where neither method moves inward noticeably.

\begin{figure}
    \centering
    \includegraphics[width=0.8\linewidth]{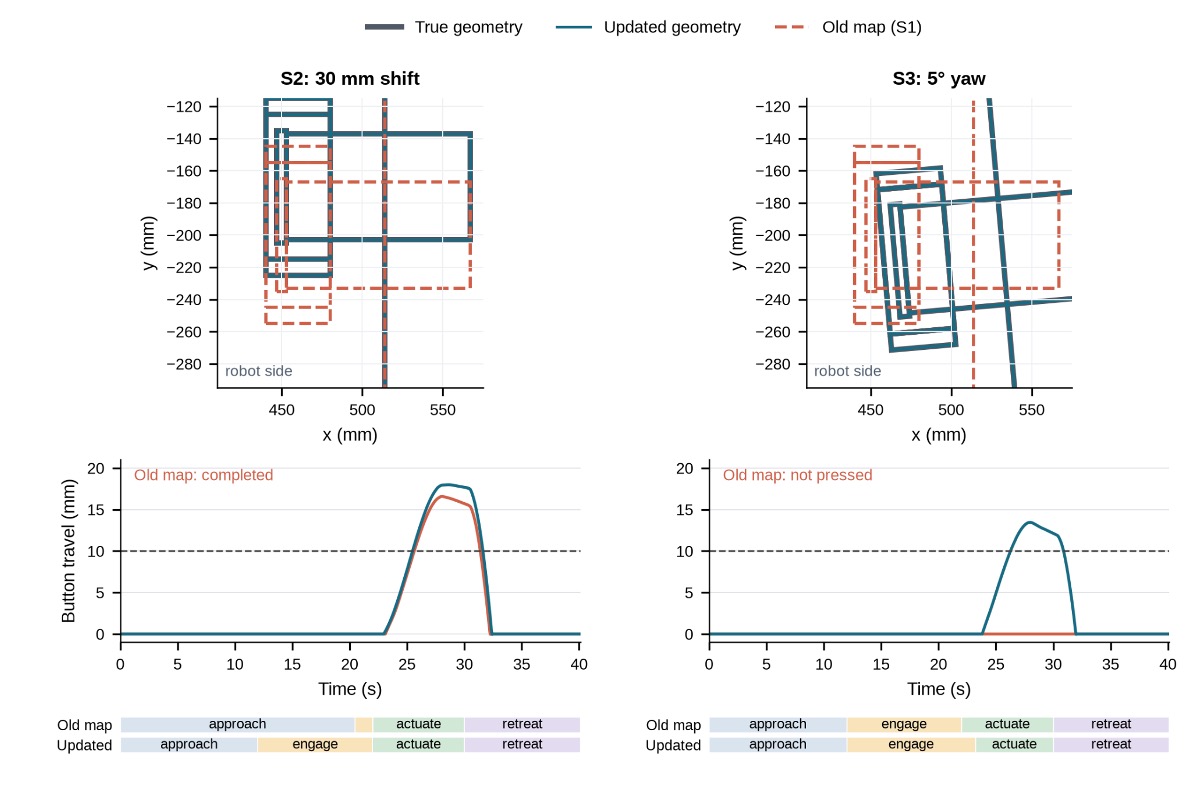}
    \caption{Effect of an outdated map after the layout changes, with the complete method (M3). The cabinet panel has moved from its nominal pose S1, either shifted by 30~mm (left, S2) or rotated by 5$^{\circ}$ (right, S3). The filter receives either the old S1 map, which no longer matches the scene, or updated geometry of the moved panel. Top: top-down view of the panel and target, showing their true pose (grey), the updated geometry (blue, coinciding with the truth) and the old map (red dashed); the robot is at the bottom. Bottom: button travel in each run, with the 10~mm trigger dashed, and the phases each run accepts.}

    \label{fig:sim-geometry}

\end{figure}

Fig.~\ref{fig:sim-geometry} shows whether M3 needs geometry that matches the current layout. In four further runs, M3 uses either the old S1 map or the updated geometry of the changed layout, here equal to the simulator truth. Under the 30~mm shift, both maps complete safely, with 16.57~mm of travel on the old map and 17.98~mm on the updated geometry. Under the 5$^{\circ}$ yaw, the old map passes through every phase but never presses the button, so the run stays incomplete, whereas the updated geometry completes safely with 13.45~mm of travel. None of the four runs has non-permitted contact. In this test, outdated geometry therefore causes a missed operation rather than unsafe contact. A 30~mm shift is tolerated, but a 5$^{\circ}$ rotation is enough to prevent the press, so the geometry used by the filter should be updated whenever the layout changes.

\subsection{Real-world Results}
\label{sec:exp-real}

\begin{figure}
    \centering
    \includegraphics[width=1\linewidth]{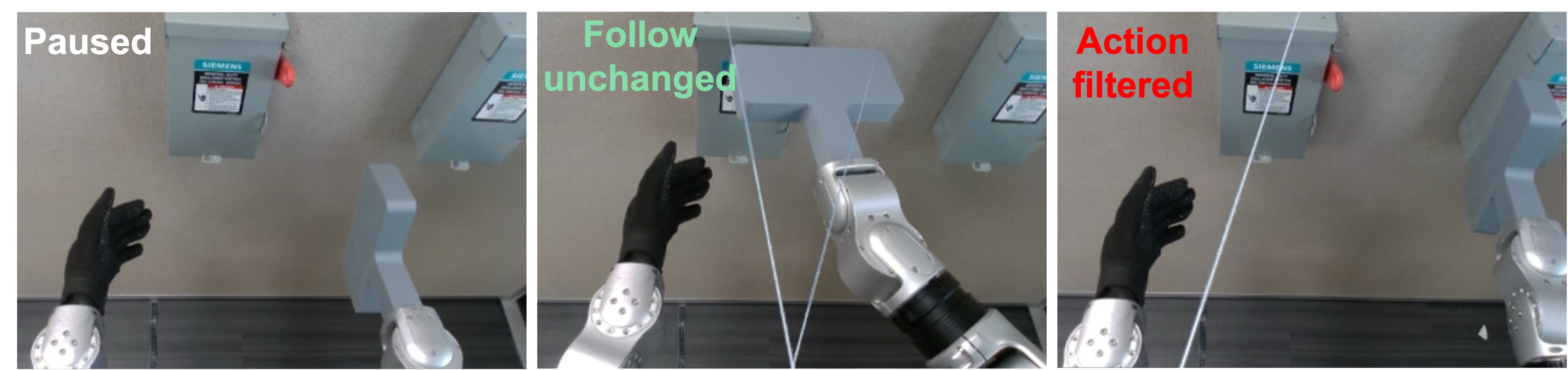}
    \caption{Operator's headset view in three filter states: paused (left), following unchanged while the tool approaches the target, the left switch (middle), and filtered while it approaches the non-target, the right switch (right).}

    \label{fig:operator-views}

\end{figure}

Table~\ref{tab:hw-results} lists all 12 attempts. In each attempt, the operator first operates the target and then deliberately approaches the non-target, so each attempt tests both functions of the filter: keeping the target reachable and stopping motion towards the non-target (Fig.~\ref{fig:operator-views}). There is no non-target actuation and no accidental contact in any of the 12 attempts, and the approach triggers an automatic return in 11 of them. Both incomplete attempts are round 6, in which the target is changed to the other switch during the operation. In C1 round 6, the tool approaches left A without fully touching it, and the filter is not modifying the command at closest approach. In C2 round 6, right B is out of reach although the tool gets near it.

\textbf{Filter intervention.} Our filter modifies the operator's command in 8.45\% of eligible cycles in C1 and 26.78\% in C2. Position, target allocation and time spent near the boundary all vary between the configurations, so these numbers describe how often the filter acts rather than how hard the task is.

\begin{figure}
    \centering
    \includegraphics[width=1\linewidth]{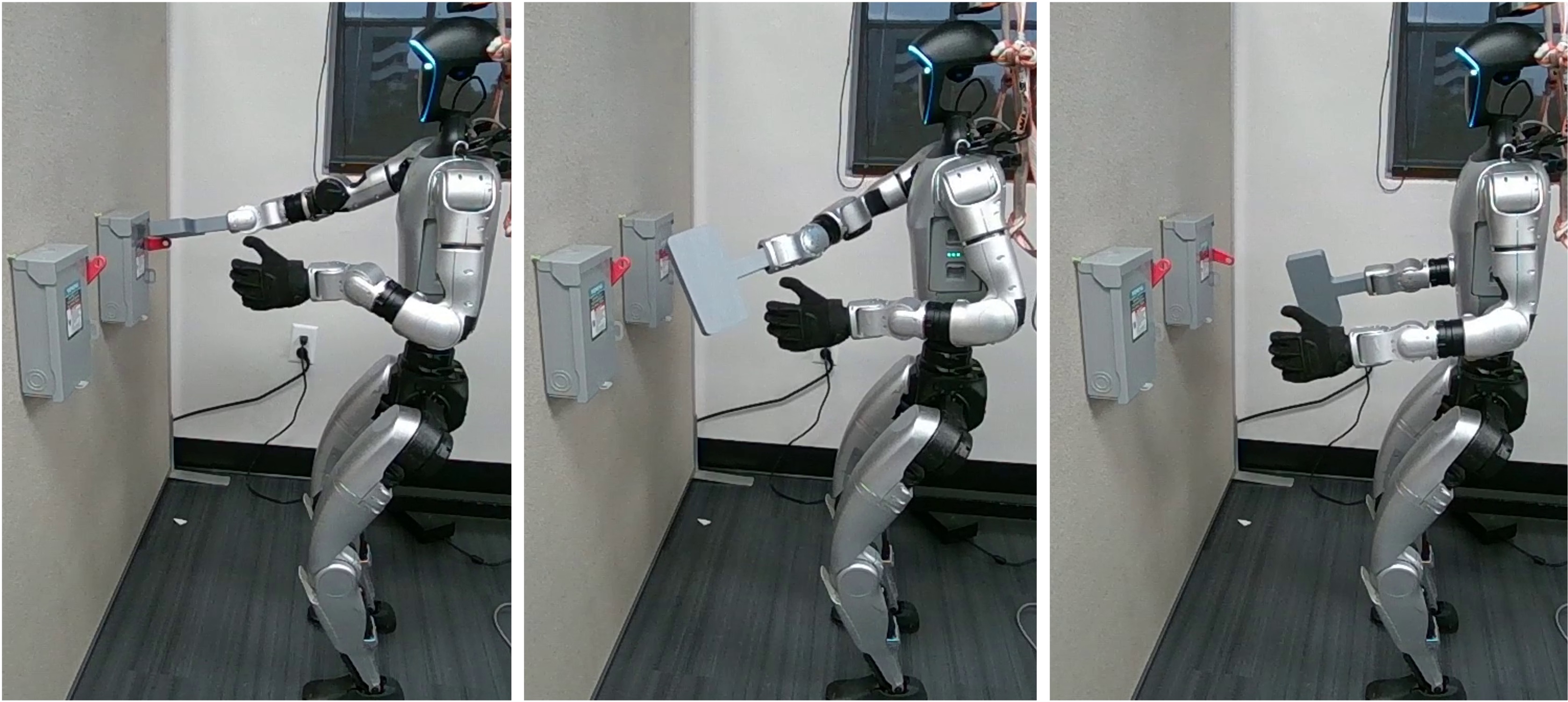}
\caption{External view of an attempt with the right switch B as target. Left: the tool operates the target switch. Middle: the tool approaches the non-target, the left switch A, and is stopped short of it. Right: the arm after the automatic return to its initial pose.}

    \label{fig:hw-sequence}
\end{figure}

\textbf{Returns and model gaps.} The auto-return column shows whether approaching the non-target triggers an automatic return. It does so in 11 of 12 rounds. A return starts when the model gap, the distance in the scene model between the tool head and the nearest constrained box, reaches $d = 50$~mm (Section~\ref{sec:method-return}), and each first automatic return begins at an encoder gap of 49.04-49.97~mm (Fig.~\ref{fig:hw-sequence}). C2 round 1 is the only exception, as explained below. The reference gap, computed at the commanded pose, never falls below 50.00~mm. The encoder gap, computed at the measured pose, reaches a minimum of 49.04~mm in C1 and 49.55~mm in C2. Each configuration has 10 encoder samples below 50~mm, and all 20 occur at the onset of a return, consistent with tracking lag between command and encoder.

\textbf{Constraint holds.} The last column in Table~\ref{tab:hw-results} gives the longest constraint hold in each round. Of 158 holds in total, 148 last at most two cycles, and the longest hold is at most 0.50~s in 10 of the 12 rounds. The longest hold, 61.38~s in C2 round 1, occurs after the tool has moved past the edge of the non-target model box, where the closest separating direction no longer supports both complete boxes. The geometry check rejects this state for 3,069 cycles, so the rows cannot be built, and the arm holds until the operator requests a manual return. An offline replay of the logged poses through the same geometry code reproduces the condition. The second-longest hold, 8.74~s in C1 round 4, occurs when the nonlinear check rejects all four step scales for 361 cycles. No QP is running during either hold, so neither reflects solver time. The first hold shows a weakness of the implementation: a hold has no timeout or automatic recovery.

\textbf{Computation.} All 180 background QP calls succeed, with a mean solve time of about 0.5~ms and a maximum of 2.49~ms. Of the 22,516 in-trial cycles that produce a candidate, 71.4\% use the nominal command, 18.6\% the active-set solution and 10.1\% the native-bound projection. Filter and dispatch time averages 5.5-5.7~ms per configuration, with a maximum of 22.2~ms. Loop work averages about 9~ms, with a maximum of 36.8~ms, and exceeds the 20~ms nominal period in 417 samples in C1 and 44 in C2, so 50~Hz is a nominal rather than a guaranteed rate. These numbers show that the non-blocking schedule of Section~\ref{sec:method-scheduling} works as intended. Most cycles need no solver at all, every candidate comes from one of the three fast paths, and background QP results enter only through the active-set step. A QP solve takes about 0.5~ms on average, a small fraction of the 20~ms period, so solver time does not limit the loop rate.

\begin{table}[width=.75\linewidth,cols=5,pos=t]
\caption{Real-world results over 12 attempts for two configurations C1 and C2.}
\label{tab:hw-results}
\begin{tabular*}{\tblwidth}{@{} LCCCCC @{}}
\toprule
Config. & Round & Target & Task done & Auto return & Longest hold (s) \\
\midrule
C1 & 1 & B & Yes & Yes & 0.02 \\
C1 & 2 & B & Yes & Yes & 0.03 \\
C1 & 3 & B & Yes & Yes & 0.50 \\
C1 & 4 & B & Yes & Yes & 8.74 \\
C1 & 5 & B & Yes & Yes & 0.24 \\
C1 & 6 & A & No  & Yes & 0.07 \\
\midrule
C2 & 1 & A & Yes & No  & 61.38 \\
C2 & 2 & A & Yes & Yes & 0.04 \\
C2 & 3 & A & Yes & Yes & 0.07 \\
C2 & 4 & A & Yes & Yes & 0.04 \\
C2 & 5 & A & Yes & Yes & 0.02 \\
C2 & 6 & B & No  & Yes & 0.07 \\
\bottomrule
\end{tabular*}
\end{table}

\section{Discussion}
\label{sec:discussion}

\subsection{Influence of the standing position on filter activity and target reach}
\label{sec:disc-position}

In practical switching operations, a humanoid cannot always stand at the exact same place in front of the equipment every time, and its position relative to the cabinets changes the geometry that the filter works with. The two configurations show this effect. The filter changes the operator's command in 26.78\% of eligible cycles in C2, compared with 8.45\% in C1. A possible reason is the distance between the working region and the non-target box. At the return pose, the tool is 28.49~cm from the non-target box in C1 but only 12.30~cm in C2, and C2 places the robot closer to the wall, about 41~cm from toe to wall compared with about 48.5~cm. The tool therefore starts each motion closer to the boundary in C2 and spends more time where the rows are active. The timing of the corrections is consistent with this explanation. In C1, no cycle is repaired before closest approach to the target, whereas in C2, rounds 2, 3 and 4 have 46, 583 and 4 repaired cycles before it, and rounds 3 and 4 also have automatic returns before it. 

The standing position also determines whether the target can be reached at all. Both incomplete operations occur in round 6, when the target is changed to the other switch. The constraints follow the change, since approaching the former target triggers an automatic return in both rounds, but the new target is not operated, and the operator attributes this to the reach of the single right-arm tool from that position. The position also affects whether a return is effective. At one candidate position for C2, the return pose lies 4.92~cm from the non-target box, inside the 5~cm boundary, so the arm returns to a pose that is itself at the boundary and approaching the non-target produces no visible retreat. Moving about 18.6~cm back restores the retreat but puts the left target out of reach. For a tool mounted on one arm, the position therefore trades target reach against return clearance. In summary, the standing position affects how often the filter acts, whether the target can be reached and whether the return is effective. The 10 of 12 reported operations reflect the reach of the tool from each position as well as the behavior of the filter.

\subsection{Causes of constraint holds and computational cost}
\label{sec:disc-holds}

A filter that runs inside teleoperation must not stall the robot or delay its commands, so the duration of constraint holds and the cost of computation determine whether the method is usable during live operation. Most holds are short: 148 of 158 last at most two cycles. A likely reason is that a short hold occurs when a candidate narrowly fails the next-pose check, and the following cycle, with a slightly different state, yields an acceptable candidate. The two long holds have different causes. The 61.38~s hold in C2 round 1 occurs after the tool has moved past the edge of the non-target box. There, the closest separating direction no longer supports both complete boxes, so the geometry check rejects the state for 3,069 cycles and no rows can be built. Because the arm does not move while held, the state does not change, and the hold persists until the operator requests a manual return. This also explains why C2 round 1 is the only round without an automatic return. The 8.74~s hold in C1 round 4 occurs when the nonlinear check rejects all four step scales for 361 cycles, which suggests that the linearized rows of Eq.~(\ref{eq:rows}) permit steps that the exact barrier does not.

Neither long hold involves a running QP, and solver time is small throughout. All 180 background QP calls succeed, with a mean solve time of about 0.5~ms, a small fraction of the 20~ms period. The QP is rarely needed: 71.4\% of the cycles that produce a candidate use the nominal command, which is expected when the tool is far from every constrained box for most of the task, and the remaining cycles are served by the active-set solution (18.6\%) and the native-bound projection (10.1\%). In summary, the non-blocking schedule keeps solver time from limiting the loop, and the main weakness of the current implementation is recovery from unusual geometric states rather than solver speed.

\subsection{Difference between commanded and measured model gaps}
\label{sec:disc-gaps}

The filter acts on the command sent to the robot, whereas safety depends on where the arm actually is. We therefore compare two model gaps to the non-target: the reference gap, computed at the commanded pose, and the encoder gap, computed at the pose measured by the joint encoders. Over all rounds, the reference gap never falls below 50.00~mm, whereas the encoder gap reaches a minimum of 49.04~mm in C1 and 49.55~mm in C2. Each configuration has 10 encoder samples below 50~mm, and all 20 occur at the onset of a return.

\begin{figure}
    \centering
    \includegraphics[width=1\linewidth]{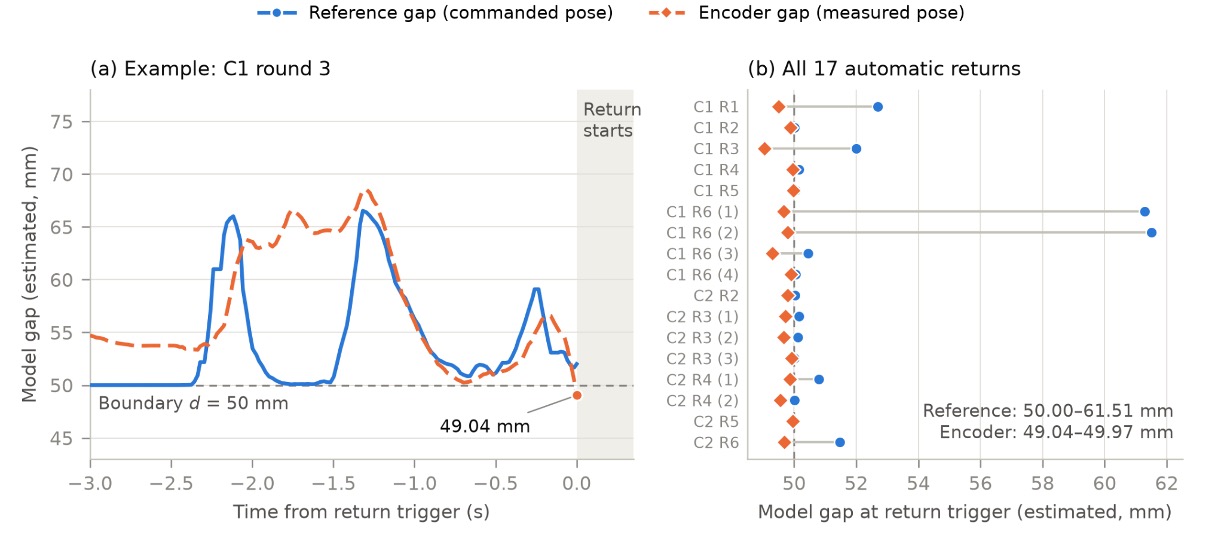}
    \caption{Reference and encoder model gaps to the non-target at automatic returns. (a) The last 3~s before the return in C1 round 3, the return with the smallest encoder gap. (b) Both gaps at the trigger cycle of each of the 17 automatic returns.}

    \label{fig:gap-lag}
\end{figure}

Fig.~\ref{fig:gap-lag} shows how these values arise. Fig.~\ref{fig:gap-lag}(a) follows the last 3~s before the return in C1 round 3, the return with the smallest encoder gap. The two gaps have nearly the same shape, but the encoder gap follows the reference gap with a short delay. In several intervals, the reference gap stays flat at 50~mm, because the filter holds the command at the boundary. Just before the trigger, the command stops outside the boundary, while the measured arm, which lags behind, continues briefly towards the non-target and reaches 49.04~mm. The automatic return is triggered at this moment. Fig.~\ref{fig:gap-lag}(b) shows both gaps at the trigger cycle of all 17 automatic returns. The reference gap is at or above 50.00~mm in every case, whereas the encoder gap is 49.04-49.97~mm, so the measured arm reaches the boundary at every trigger but crosses it by less than 1~mm. In most returns, the two gaps nearly coincide. In the first two returns of C1 round 6, however, the reference gap is still about 61~mm when the encoder gap reaches the boundary, so these returns are triggered by the measured pose. This confirms that the return logic of Section~\ref{sec:method-return} checks both poses, so a lag between them cannot delay the return.

One possible reason for the brief crossing is therefore tracking lag between the command and the arm, rather than a failure of the constraint. In summary, the constraint holds exactly on the command and approximately on the measured pose, and the margin $d$ must cover tracking lag as well as model error. With $d = 5$~cm, the observed crossing of less than 1~mm uses a small part of the margin.

\subsection{Effects of part-region permission and phase-direction constraints}
\label{sec:disc-sim}

Switch operation requires contact with the target, so a filter must decide which contact is allowed and how the hand may move while in contact. Our simulation study separates the contributions of the two extensions that address these questions. Under I1, a finger pivot brings the middle finger towards the target. M1 removes every pair between the robot and the target assembly once permission is active, so no row prevents the middle finger pad from touching the guard, which it does with peak summed forces of 8.00-9.35~N for 5.18-5.45~s per run. M3 keeps a 5~mm clearance for every right-hand part except the index pad, so the same input is corrected before contact. Because the index pad pair is permitted in both methods, the press itself is unaffected, and both methods complete 3/3 runs. Object-level permission therefore admits contact by any part of the hand once the target is excluded, which is the permission that the hardware filter uses.

Removing the phase-direction rows has different effects under the two disturbed inputs. Under I1, M2 accepts engage but never actuate. In S1 and S2, the QP reports no solution for 932 and 944 cycles after engage, and the arm holds. In S3, the actuate request stays pending for 400 cycles because the index tip is not near the front face of the button, and the run passes to retreat without a press. A possible reason is that, without the centering corridor, the pivot moves the finger away from the button axis, so the finger either reaches a state where the clearances and the other rows cannot be satisfied together, or never meets the actuate gate. Under I2, M2 completes all three runs safely, and the phase-direction rows change only the motion. The corridor of Eq.~(\ref{eq:corridor}) bounds the lateral speed towards the gate, which explains why the M3 lateral path stays at 45.28-49.67~mm, compared with 114.53-117.32~mm for M2. The retreat rows forbid inward motion, which removes the inward motion of about 3.4~mm seen with M2 in S1 and S2. In S3, M2 moves inward by only 0.016~mm, so the rows have little to remove. In summary, part-region permission prevents contact by unintended parts of the hand, and phase-direction constraints keep the finger centered during contact and outward during withdrawal.

\subsection{Relation between the simulation and real-world results}
\label{sec:disc-sim-real}

Table~\ref{tab:sim-real} compares the two studies. They share the robot model and the filter formulation, but differ in task, end effector and source of geometry, so they complement each other rather than repeat the same measurement.

\begin{table}
\caption{Comparison of our simulation and real-world studies.}
\label{tab:sim-real}
\begin{tabular*}{\tblwidth}{@{\extracolsep{\fill}} lll @{}}
\toprule
 & Simulation & Real world \\
\midrule
Task & Press a button & Operate a handle \\
End effector & Three-finger hand & Rigid wrist tool \\
Geometry & Simulator truth & Head RGB-D estimate \\
Trials & 3 layouts $\times$ 3 inputs & 2 positions $\times$ 6 rounds \\
Outcome & 9/9 safe (M3), 5/9 (M0) & No non-target actuation \\
Holds & 0 (M3) & 158 (148 $\le$ 2 cycles) \\
QP time & $\le$ 0.747~ms & $\approx$ 0.5~ms (mean) \\
Geometry change & 5$^{\circ}$ map error: no press & Intervention 8.45\% vs 26.78\% \\
\bottomrule
\end{tabular*}
\end{table}

Three findings are consistent across the two studies. First, the QP does not limit the control loop, as its solve time stays well below 1~ms in both. Second, no unintended contact occurs when the filter's geometry matches the scene. Third, the position of the equipment relative to the robot can change the filter's behavior: a different standing position more than triples the hardware intervention ratio, and an unreflected 5$^{\circ}$ rotation prevents the press in simulation.

The studies differ in constraint holds, which occur only on hardware. A possible reason is that the simulation uses ground-truth geometry and joint states, whereas on hardware the boxes are estimated, the measured pose lags the reference and operator motion varies between rounds, creating states near the boundary that the simulation runs do not reach. The simulation thus isolates the effect of each constraint, while the hardware study shows how the complete system behaves under real perception and operator input.

\subsection{Implications for real-world electrical switch operations}
\label{sec:disc-implications}

Switching operations follow procedures that identify the target device, operate it and then verify the result. The results suggest three implications for introducing a teleoperated humanoid into this workflow. First, the filter addresses the physical operation step, while target identification and verification remain with the operator and the procedure. The operator selects the target before each trial, and the filter then keeps that target reachable and constrains its neighbor. Across the 12 attempts, the operator reports no non-target actuation, and approaching the non-target triggers an automatic return in 11 rounds. A safety layer of this kind can therefore reduce the reliance on the operator's precision near a neighboring device, but it does not replace the procedural checks around the operation.

Second, the method fits sites that are designed for human workers. The scene model is built from the robot's own head camera at its current position without entering a cabinet pose for each scene or modifying the equipment. This matters in electrical rooms, where devices sit close together and drawings may not match the installed layout. The simulation adds that the model should be rebuilt when the robot or the equipment moves, since an outdated map causes a missed operation after a 5$^{\circ}$ rotation.

Third, the method keeps authority with the operator and fails towards stopping. Safe commands pass through unchanged, a stop at the boundary can only be lifted through recalibration and an explicit command, and when the rows cannot be built, the arm holds instead of moving. For a site procedure, this means that the operator must be prepared to resolve a hold, for example by requesting a manual return, until a fallback is implemented. The simulation further suggests that, for tasks in which the hand itself presses a device, permissions should be limited to the intended contact and motion should be restricted by task phase.

\subsection{Limitations}
\label{sec:limitations}

Several limitations remain. First, the hardware and simulation studies use different end effectors, so the fine contact control examined in simulation is not tested on the robot. In simulation, the robot presses the button with the index finger of a three-finger hand, which makes it possible to permit contact for one finger pad and to protect the other fingers. On our physical robot, the switch is operated with a rigid 3D-printed tool on the right wrist, which the filter treats as a single box. The hardware study therefore tests only object-level permission, with the whole target assembly excluded from the constraints, and cannot show whether part-region permission and phase-direction constraints would work with a real hand. The tool is also mounted on one arm only, which limits the reach of the robot and causes both incomplete operations.

Second, our physical test equipment differs from that of real electrical substations. The robot operates two general-duty enclosed safety switches mounted on a wall, of the kind used in residential and light commercial installations. Substation switchgear, such as disconnect switches, circuit breakers and racking mechanisms, can differ in size, handle design, operating force and travel, and is usually housed in large cabinets or arranged in rows of panels. We do not test such cabinets, so the scene model, the declared size bound and the standing positions used here may not transfer directly, and larger or more closely packed equipment may require other box sizes, more constrained objects and other positions.

Third, the switches are not energized. Energizing them is not feasible in our laboratory, and testing an unproven filter on live equipment would be unsafe. As a result, we cannot measure the electrical consequences of an operation, such as arcing during switching or the effect of an incomplete or repeated actuation, and the study provides no evidence for live work. The results describe the motion of the robot relative to de-energized equipment only. 
For the same reason, this study does not assess how an arc or the electromagnetic field near live equipment would affect the robot. An arc flash releases intense heat, pressure and light, and the cameras, encoders, motor drivers and onboard computers of a humanoid can be damaged or disturbed, which can in turn degrade the scene model or the control loop on which the filter depends. Damage to a robot is less serious than injury to a human worker, but a robot that fails next to electrical equipment can still create a hazard. Before industrial deployment, the robot would therefore need its own protection, such as insulated tools, arc-rated covers, an adequate standoff distance and verified electromagnetic compatibility. We leave this part to future work.

\section{Conclusion}
This study proposes a runtime safety filter for VR teleoperation of a humanoid robot on neighboring safety switches. Humanoid robots can perform switching operations at sites designed for human workers, but in a tight workspace an imperfect operator command can bring the tool into a neighboring device, and existing humanoid teleoperation systems do not constrain motion against the scene. A safety layer for this task must keep the selected target reachable while stopping motion towards its neighbor, work from the robot's own sensing at an unmodified site, and leave safe commands unchanged. To address these challenges, a digital model of the scene is built from the robot's head RGB-D camera at its current standing position, using fitted cabinet planes and a declared size bound. Based on the operator-selected target, a CBF-QP filter keeps the tool at least 5~cm from the non-target assembly and the walls while leaving the target assembly unconstrained. A non-blocking schedule passes safe commands unchanged without delaying the control loop, and a latched return, which only the operator can lift, stops continued inward motion at the boundary. On a Unitree G1, in 12 attempts at two standing positions, there are 10 completed target operations and no non-target actuation or accidental contact. Approaching the non-target triggers an automatic return in 11 rounds, the reference model gap never falls below 50~mm, and all 180 background QP calls succeed with a mean solve time of about 0.5~ms. In our simulation study, part-region permissions and phase-direction constraints together achieve 9/9 safe completions, compared with 5/9 for raw teleoperation: the former removes non-permitted contact by other parts of the hand, and the latter keeps the finger centered and prevents it from pushing back in during withdrawal. This work demonstrates the feasibility of adding a target-aware safety layer to live humanoid control next to neighboring electrical switches, without modifying the site or the existing software stack, and provides a technical foundation for using humanoid robots in electrical switching operations.

Although this study demonstrates the approach on de-energized equipment in a laboratory, several challenges remain before it can be applied to real switching operations. First, the robot operates the switches with a rigid tool on one arm, whereas the fine contact control is examined only in simulation with a three-finger hand, so part-region permission and phase-direction constraints have not been tested on the robot, and the single tool also limits reach. Second, the test equipment consists of wall-mounted general-duty safety switches, which may differ from substation switchgear in size, handle design, operating force and cabinet arrangement. Third, the switches are not energized, so the electrical consequences of an operation and the effect of arcing on the robot itself cannot be assessed. 
Future research will therefore focus on bringing the part-region and phase logic to a dexterous hand on the robot, testing on substation-type switchgear and large cabinets, and evaluating energized operation under appropriate electrical safety arrangements.

\section*{CRediT authorship contribution statement}
Songyang Liu: Writing – original draft, Investigation, Software, Visualization, Methodology. Yu Hou: Supervision. Yuqing Hu: Supervision. Shuai Li: Writing – review \& editing, Validation, Supervision, Conceptualization, Resources, Project administration.
\section*{Declaration of competing interest}
The authors declare that they have no known competing financial
interests or personal relationships that could have appeared to influence
the work reported in this paper.

\section*{Declaration of generative AI use}
None.

\section*{Data availability}
Data will be made available on request.

\section*{Acknowledgements}
This research was supported by the U.S. National Science Foundation (No. 2222810). Any
opinions, findings, and conclusions or recommendations
expressed in this paper are those of the authors and do
not necessarily reflect the views of the National Science
Foundation, the University of Florida.

\bibliographystyle{cas-model2-names}
\bibliography{references}

@misc{dvids2022cemirt,
  author       = {Moeder, Brooke},
  title        = {31st {CES}, {CEMIRT}, supply power to 31 {FW}},
  year         = {2022},
  howpublished = {Defense Visual Information Distribution Service (DVIDS), U.S. Air Force photo, VIRIN 220621-F-ZR251-1040},
  url          = {https://www.dvidshub.net/image/7338359/31st-ces-cemirt-supply-power-31-fw},
  note         = {Public domain. Accessed 12 November 2026}
}

@InProceedings{allan2014substations,
  author    = {Allan, Jean-Francois and Beaudry, Julien},
  title     = {Robotic Systems Applied to Power Substations -- A State-of-the-Art Survey},
  booktitle = {Proceedings of the 3rd International Conference on Applied Robotics for the Power Industry},
  year      = {2014},
  pages     = {1--6},
  doi       = {10.1109/CARPI.2014.7030049},
}

@Article{lu2017substation,
  author  = {Lu, Shouyin and Zhang, Ying and Su, Jianjun},
  title   = {Mobile Robot for Power Substation Inspection: A Survey},
  journal = {IEEE/CAA Journal of Automatica Sinica},
  year    = {2017},
  volume  = {4},
  number  = {4},
  pages   = {830--847},
  doi     = {10.1109/JAS.2017.7510364},
}

@Misc{cigre2020robotics,
  author       = {{CIGRE Working Group B3.47}},
  title        = {Application of Robotics in Substations},
  year         = {2020},
  howpublished = {CIGRE Technical Brochure 807},
}

@Article{lin2024grounding,
  author  = {Lin, Yunhan and Wang, Jiahui and Liu, Kaibo and Min, Huasong},
  title   = {Grounding Rod Hanging and Removing Robot with Hand-Eye Self-Calibration Capability in Substation},
  journal = {Complex \& Intelligent Systems},
  year    = {2024},
  volume  = {10},
  number  = {5},
  pages   = {6491--6507},
  doi     = {10.1007/s40747-024-01492-2},
}

@InProceedings{jia2025switchgear,
  author    = {Jia, Zhaoxin and Wang, Hailei and Guo, Rui and Bai, Jinhui and Yang, Xinyu and Wang, Zheng},
  title     = {End-to-End Learning-Based Autonomous Operation Robot for Substation Switchgear},
  booktitle = {Proceedings of the 2nd International Conference on Smart Grid and Artificial Intelligence},
  year      = {2025},
  pages     = {1439--1443},
  doi       = {10.1109/SGAI64825.2025.11009698},
}

@Misc{zhao2025teleoperation,
  author       = {Zhao, Yuxuan and Tang, Yuanchen and Zhang, Jindi and Yu, Hongyu},
  title        = {Empowering Robot Teleoperation: Exploring the Synergies Between Devices and Manipulator Controllers in a Comparative Study},
  year         = {2025},
  howpublished = {arXiv:2511.07720},
}

@InProceedings{cheng2024opentelevision,
  author    = {Cheng, Xuxin and Li, Jialong and Yang, Shiqi and Yang, Ge and Wang, Xiaolong},
  title     = {Open-{TeleVision}: Teleoperation with Immersive Active Visual Feedback},
  booktitle = {Proceedings of the Conference on Robot Learning},
  year      = {2024},
  note      = {arXiv:2407.01512},
}

@InProceedings{he2024omnih2o,
  author    = {He, Tairan and Luo, Zhengyi and He, Xialin and Xiao, Wenli and Zhang, Chong and Zhang, Weinan and Kitani, Kris and Liu, Changliu and Shi, Guanya},
  title     = {{OmniH2O}: Universal and Dexterous Human-to-Humanoid Whole-Body Teleoperation and Learning},
  booktitle = {Proceedings of the Conference on Robot Learning},
  year      = {2024},
  note      = {arXiv:2406.08858},
}

@Misc{ze2025twist2,
  author       = {Ze, Yanjie and Zhao, Siheng and Wang, Weizhuo and Kanazawa, Angjoo and Duan, Rocky and Abbeel, Pieter and Shi, Guanya and Wu, Jiajun and Liu, C. Karen},
  title        = {{TWIST2}: Scalable, Portable, and Holistic Humanoid Data Collection System},
  year         = {2025},
  howpublished = {arXiv:2511.02832},
}

@Article{luo2026sonic,
  author  = {Luo, Zhengyi and Yuan, Ye and Wang, Tingwu and Li, Chenran and Casta{\~n}eda, Fernando and Chen, Sirui and Cao, Zi-Ang and Li, Jiefeng and Minor, David and Ben, Qingwei and Park, Jinhyung and Sami, David and Wang, Zi and Da, Xingye and Ding, Runyu and Hogg, Cyrus and Song, Lina and Lim, Edy and Jeong, Eugene and He, Tairan and Xue, Haoru and Xiao, Wenli and Yuen, Simon and Kautz, Jan and Chang, Yan and Iqbal, Umar and Fan, Linxi and Zhu, Yuke},
  title   = {{SONIC}: Supersizing Motion Tracking for Natural Humanoid Whole-Body Control},
  journal = {Science Robotics},
  year    = {2026},
  volume  = {11},
  number  = {117},
  pages   = {eaed4592},
  doi     = {10.1126/scirobotics.aed4592},
}

@InProceedings{ames2019cbf,
  author    = {Ames, Aaron D. and Coogan, Samuel and Egerstedt, Magnus and Notomista, Gennaro and Sreenath, Koushil and Tabuada, Paulo},
  title     = {Control Barrier Functions: Theory and Applications},
  booktitle = {Proceedings of the 18th European Control Conference},
  year      = {2019},
  pages     = {3420--3431},
  doi       = {10.23919/ECC.2019.8796030},
}

@Misc{chen2025pssa,
  author       = {Chen, Rui and Sun, Yifan and Liu, Changliu},
  title        = {Dexterous Safe Control for Humanoids in Cluttered Environments via Projected Safe Set Algorithm},
  year         = {2025},
  howpublished = {arXiv:2502.02858},
}

@InProceedings{selvaggio2016vision,
  author    = {Selvaggio, Mario and Chen, Fei and Gao, Boyang and Notomista, Gennaro and Trapani, Francesco and Caldwell, Darwin},
  title     = {Vision Based Virtual Fixture Generation for Teleoperated Robotic Manipulation},
  booktitle = {Proceedings of the International Conference on Advanced Robotics and Mechatronics},
  year      = {2016},
  pages     = {190--195},
  doi       = {10.1109/ICARM.2016.7606917},
}

@Article{brunke2025semantic,
  author  = {Brunke, Lukas and Zhang, Yanni and R{\"o}mer, Ralf and Naimer, Jack and Staykov, Nikola and Zhou, Siqi and Schoellig, Angela P.},
  title   = {Semantically Safe Robot Manipulation: From Semantic Scene Understanding to Motion Safeguards},
  journal = {IEEE Robotics and Automation Letters},
  year    = {2025},
  volume  = {10},
  number  = {5},
  pages   = {4810--4817},
}

@Misc{kim2024armor,
  author       = {Kim, Daehwa and Srouji, Mario and Chen, Chen and Zhang, Jian},
  title        = {{ARMOR}: Egocentric Perception for Humanoid Robot Collision Avoidance and Motion Planning},
  year         = {2024},
  howpublished = {arXiv:2412.00396},
}

@Misc{gadde2026recal,
  author       = {Gadde, Mohitvishnu S. and Malik, Ashish and Dugar, Pranay and Shrestha, Aayam Kumar and Li, Fuxin and Fern, Alan},
  title        = {Collision-Aware Humanoid Whole-Body Control under Imperfect Tracking Targets},
  year         = {2026},
  howpublished = {arXiv:2609.16405v2},
}

@Misc{wilkinson2026vlpsa,
  author       = {Wilkinson, Meg and Fourney, Emily and Burdick, Joel W. and Ames, Aaron D.},
  title        = {{VLPSA}: Vision-Language-Poisson-Safe Actions for Full-Body Safety of Learned Policies},
  year         = {2026},
  howpublished = {arXiv:2609.22462},
}

@Misc{osha1910_333,
  author       = {{Occupational Safety and Health Administration}},
  title        = {1910.333 -- Selection and Use of Work Practices},
  year         = {2024},
  howpublished = {\url{https://www.osha.gov/laws-regs/regulations/standardnumber/1910/1910.333}},
  note         = {Accessed: 2026-07-02},
}

@Misc{snopud2026switching,
  author = {{Snohomish County Public Utility District No. 1}},
  title  = {Switching and Clearance Procedures Manual},
  year   = {2026},
  note   = {Revised May 15, 2026; Accessed: 2026-07-02},
}

@Misc{prpa2021switching,
  author = {{Platte River Power Authority}},
  title  = {{SOP} \#1: Power System Switching Procedure},
  year   = {2021},
  note   = {Version 21; Accessed: 2026-07-02},
}

@Article{yang2025substation,
  author  = {Yang, Q. and Dong, J. and Tan, M. and Wang, J. and Guo, D. and Kang, H. and Wang, P.},
  title   = {A novel navigation assistant method for substation inspection robot based on multisensory information fusion},
  journal = {Journal of Advanced Research},
  year    = {2025},
  doi     = {10.1016/j.jare.2025.01.016},
}

@Misc{avangrid2024spot,
  author       = {{Avangrid}},
  title        = {Avangrid Pilots Mobile Robot Dog to Advance Substation Inspections with {AI}},
  year         = {2024},
  howpublished = {\url{https://www.avangrid.com/w/avangrid-pilots-mobile-robot-dog-to-advance-substation-inspections-with-ai}},
  note         = {Accessed: 2026-07-02},
}

@Misc{morton2026constrainedmimic,
  author       = {Morton, Daniel and Mohnot, Pranit and Pavone, Marco},
  title        = {Constrained Whole-Body Tracking for Humanoid Robots},
  year         = {2026},
  howpublished = {arXiv:2606.00374v2},
  note         = {To be presented at the 2026 Conference on Robot Learning},
}

@Misc{zhao2025xrobotoolkit,
  author       = {Zhao, Zhigen and Yu, Liuchuan and Jing, Ke and Yang, Ning},
  title        = {{XRoboToolkit}: A Cross-Platform Framework for Robot Teleoperation},
  year         = {2025},
  howpublished = {arXiv:2508.00097},
}

@Article{arnstrom2022daqp,
  author  = {Arnstr{\"o}m, Daniel and Bemporad, Alberto and Axehill, Daniel},
  title   = {A Dual Active-Set Solver for Embedded Quadratic Programming Using Recursive {LDL}$^{T}$ Updates},
  journal = {IEEE Transactions on Automatic Control},
  year    = {2022},
  volume  = {67},
  number  = {8},
  pages   = {4362--4369},
  doi     = {10.1109/TAC.2022.3176430},
}

\end{document}